\documentclass[final,5p,times,twocolumn]{elsarticle}

\usepackage{graphics}
\usepackage{graphicx}
\usepackage{epsfig}

\usepackage{amssymb}
\usepackage{amsthm}
\usepackage{amsmath}

\usepackage{lineno}
\usepackage{mathptmx}      
\usepackage{ifthen}
\usepackage{subfig}
\usepackage{epstopdf}
\usepackage{url}
\usepackage[ruled,lined,titlenumbered,linesnumbered]{algorithm2e}
\usepackage{hyperref}
\usepackage{multicol}
\usepackage{multirow}

\journal{Computers \& Operations Research}

\begin{document}

\begin{frontmatter}



\title{Memetic collaborative approaches for finding balanced incomplete block designs
\tnoteref{agradece}}
\tnotetext[agradece]{This work is partially funded by Junta de Andaluc\'{\i}a (project P10-TIC-6083, DNEMESIS -- \url{http://dnemesis.lcc.uma.es/wordpress/}), Ministerio Espa\~{n}ol de Econom\'{\i}a y Competitividad (projects TIN2014-56494-C4-1-P, UMA::EPHEMECH -- \url{https://ephemech.wordpress.com/} and TIN2017-85727-C4-1-P, UMA::DeepBio
	-- \url{http://deepbio.wordpress.com}), and Universidad de M\'{a}laga, Campus de Excelencia Internacional Andaluc\'{i}a Tech.}




\author[A]{David Rodr\'{i}guez Rueda}
\ead{drodri@unet.edu.ve}

\author[B]{Carlos Cotta}
\ead{ccottap@lcc.uma.es}

\author[B]{	Antonio J. Fern\'{a}ndez-Leiva}
\ead{afdez@lcc.uma.es}

\address[A]{Universidad Nacional Experimental del T\'{a}chira (UNET), Laboratorio de Computaci\'{o}n de Alto Rendimiento (LCAR), San Crist\'{o}bal, T\'{a}chira, 5001, Venezuela}
\address[B]{Universidad de M\'{a}laga, ETSI Inform\'{a}tica, Campus de Teatinos, 29071 M\'{a}laga, Spain}

\begin{abstract}
The balanced incomplete block design (BIBD) problem is a difficult combinatorial problem with a large number of symmetries, which add complexity to its resolution.
In this paper, we propose a dual (integer) problem representation that serves as an alternative to the classical binary formulation of the problem. We attack this problem  incrementally: firstly, we propose basic algorithms (i.e. local search techniques and genetic algorithms) intended to work separately on the two different search spaces (i.e. binary and integer); secondly,  we propose two hybrid schemes: an integrative approach (i.e. a memetic algorithm) and a collaborative model in which the previous methods work in parallel, occasionally exchanging information. 
Three distinct two-dimensional structures are proposed as communication topology among the algorithms involved in the collaborative model, as well as a number of migration and acceptance criteria for sending and receiving data. An empirical analysis comparing a large number of instances of our schemes (with algorithms possibly working on different search spaces and with/without symmetry breaking methods)
 shows that some of these algorithms can be considered the state of the art of the metaheuristic methods applied to  finding BIBDs.
Moreover, our cooperative proposal is a general scheme from which distinct algorithmic variants  can be instantiated to handle symmetrical optimisation problems. For this reason, we have also analysed its key parameters, thereby providing general guidelines for the design of efficient/robust cooperative algorithms devised from our proposal. 
\end{abstract}

\end{frontmatter}


\section{Introduction}
The generation of block designs is a well-known combinatorial problem of enormous difficulty \cite{colbourn2010crc}. The problem has a number of variants \cite{PartiallyHinkelmann2005,Bruinsma2014,Buratti1999103,Ching-Shui2014,Tsubaki2008}, among which a popular one is the so-called balanced incomplete block design (BIBD). Basically, a BIBD is defined as an arrangement of $v$ different objects into $b$ blocks such that each block contains exactly $k$ different objects, each object occurs in exactly $r$ different blocks, and every two different objects occur together in exactly $\lambda$ blocks (for $k, r, \lambda > 0$). The construction of BIBDs was initially tackled in the area of experimental design \cite{Lint+:CCombinatorics1992,Mead:DoE1993};  however, nowadays BIBDs are applied in a variety of fields such as cryptography \cite{buratti:cryptography1999}, coding theory
\cite{DBLP:journals/tcom/LanTLMH08}, food evaluation \cite{dos2014balanced}, load balance in distributed networks \cite{basu2014design}, and classification tasks \cite{madsen2014new}, among others.

BIBD generation is an NP-hard problem \cite{Corneil+:algo-techniques-adm1978} that provides an excellent benchmark for optimisation algorithms since it is scalable and has a wide variety of problem
instances ranging from easy instances to very difficult ones. As discussed in Sect.~\ref{sect:related work}, complete methods (including exhaustive search) have been applied to the problem although it
remains intractable even for designs of a relatively small size \cite{gibbons-design-theory-1996}. In fact, as proof of the difficulty of the problem, there are currently a number of open instances that have not yet been solved (although, it may be that there is no solution for them; then again, non-solvability cannot be established by complete methods). The application of metaheuristics thus seems to be appropriate to tackle larger problem instances due to the limitations of complete methods. Indeed, some approaches in this area have already provided evidence of the potential of metaheuristic approaches applied to this problem, e.g. \cite{DBLP:conf/cp2003/PrestwichCP03,RuedaCF09,faghihi2014varphi,DBLP:journals/ijcopi/RuedaCL11}. 

One of the most interesting features of the BIBD is its highly-symmetrical nature. This introduces a number of considerations that have to be taken into account. Firstly, the existence of solutions that are equivalent with respect to the same  representation space generally increases the size of the search space and, as a direct consequence,  the difficulty of finding solutions (i.e. the problem solving  complexity). In the last few decades, a number of methods have been applied to deal with symmetries~\cite{benhamou1994study,backofen2002excluding,fahle2001,Gent99symmetrybreaking}. The primary method of dealing with them consists in applying some symmetry breaking technique. This method basically imposes new constraints to remove symmetries with the goal of reducing the  problem's search space. Symmetry breaking can be applied in many diverse forms \cite{DBLP:journals/ai/MeseguerT01}. In connection with this, it is also well known that the encoding of solutions can drastically affect the search process, because it influences the underlying landscape and its navigability. This paper proposes an alternative --and novel, to the best of our knowledge-- representation scheme for BIBD solutions that we call the dual (or decimal) formulation (see Sect. \ref{sec:DualModel}), in response to the `more natural' primal (or binary) model considered in the scientific literature, cf. Sect \ref{subsec:primal model}. A number of algorithms to tackle the BIBD problem are subsequently considered to take into account the large number of possible scenarios that  arise from the combination of these two different encodings, as well as the symmetry-breaking constraints for the  BIBD problem (see Sect. \ref{sec:symetriesbreaking}). Moreover, each scenario is tackled with a number of metaheuristic techniques, including local search and genetic algorithms. As a further step, this paper also proposes mechanisms for hybridising these algorithms.  In particular, we consider both an integrative model (Sect. \ref{subsect:MAs}) and a collaborative scheme (Sect. \ref{Sec:EsquemasCooperativos}). The latter, in particular, defines a network (i.e. a set) of algorithms that intensify the search in certain parts of the search space; the communication strategy among these algorithms is defined by a certain spatial structure.  Three different topologies are considered for this purpose. We also study different policies to control communication among algorithms, i.e. which information should be submitted and when/how it should be handled by the metaheuristics in the network.  The resulting techniques are exhaustively analysed from an empirical point of view in Sect. \ref{sec:experiments}. The next section provides an overview of the  problem's foundations as well as a brief look at related work.

This paper proposes a (novel) formulation for the representation of BIBDs and a number of metaheuristics (based on this 
formulation) to handle the problem. This paper also describes a large number of metaheuristic approaches to deal with the generation of BIBDs. Some of these (i.e. the cooperative methods) constitute state-of-the-art metaheuristic methods to handle the problem. Moreover, we provide a general scheme from which other (possibly cooperative) metaheuristics can be generated. Finally, we also propose a methodology to address, in a general way, symmetrical combinatorial problems so that our methods can be easily adjusted to deal with  other symmetrical combinatorial problems.

\section{Background}

This section discusses how the BIBD problem has been tackled in the literature. The formal classical formulation of the problem is also provided.

\subsection{BIBD: Formulation and primal (or binary) model}
\label{subsec:primal model}
\sloppypar As mentioned in the introduction, a BIBD can be specified with five parameters $\langle v,b,r,k,\lambda\rangle$. Using this notation, a $\langle v,b,r,k,\lambda\rangle$-BIBD 
problem consists of dividing a set of $v$ objects into $b$ subsets of $k < v$ objects each, such that each object belongs to $r$ different subsets and any pair of objects appear together in  exactly $\lambda < b$ subsets. A standard way of representing the solution to such a problem --termed here as the {\em primal (or binary) model (B)}-- is in terms of its incidence matrix $M \equiv \{m_{ij}\}_{v \times b}$, which is a $v \times b$ binary matrix where $m_{ij} \in \{0,1\}$ is equal to 1 if the $i$th object is contained in the $j$th block, and 0 otherwise; thus, it is easy to see that each row corresponds to an object and each column to a block, and that a matrix representing a feasible solution has exactly $r$ ones per row, $k$ ones per column, and the scalar product of any pair of different rows is $\lambda$. Figure~\ref{fig:instanciaBIBD} shows configurations of the incidence matrix $M$ representing possible solutions to a $\langle8,14,7,4,3\rangle-$BIBD and a symmetric (i.e. $b = v$) $\langle7,7,3,3,1\rangle-$BIBD, respectively.

\begin{figure*}[!ht]
	\begin{center}
		\subfloat[]{
				$\left[
				\begin{tabular}{cccccccccccccc}
					0 &0& 0& 1& 0& 1& 1& 1& 0& 0& 0& 1& 1& 1 \\
					1 &1& 0& 1& 1& 0& 1& 0& 0& 1& 0& 0& 0& 1 \\
					0 &1& 1& 1& 1& 1& 0& 0& 1& 0& 0& 0& 1& 0 \\
					0 &0& 0& 0& 1& 1& 1& 1& 1& 1& 1& 0& 0& 0 \\
					1 &0& 1& 0& 1& 1& 0& 0& 0& 0& 1& 1& 0& 1 \\
					0 &1& 0& 0& 0& 0& 0& 0& 1& 1& 1& 1& 1& 1 \\
					1 &1& 1& 0& 0& 0& 1& 1& 1& 0& 0& 1& 0& 0 \\
					1 &0& 1& 1& 0& 0& 0& 1& 0& 1& 1& 0& 1& 0 \\
				\end{tabular}
				\right]$
			}
		\subfloat[]{
			$\left[
			\begin{tabular}{c c c c c c c }
				0 & 1 & 0 & 1 & 0 & 1 & 0  \\
				1 & 0 & 0 & 1 & 0 & 0 & 1 \\
				1 & 1 & 1 & 0 & 0 & 0 & 0  \\
				0 & 0 & 1 & 0 & 0 & 1 & 1  \\
				0 & 1 & 0 & 0 & 1 & 0 & 1  \\
				1 & 0 & 0 & 0 & 1 & 1 & 0  \\
				0 & 0 & 1 & 1 & 1 & 0 & 0  \\
			\end{tabular}
			\right]$
		}
\end{center}
	\caption{(a) A $\langle 8,14,7,4,3\rangle-$BIBD. (b) A $\langle 7,7,3,3,1\rangle-$symmetric BIBD.} 
	\label{fig:instanciaBIBD}
\end{figure*}
\normalsize

Note that the five parameters defining a $\langle v,b,r,k,\lambda\rangle-$BIBD are interrelated and satisfy the following two relations: $bk = vr$ and $\lambda ( v - 1) = r ( k - 1)$. This means that we could define an instance using just three parameters $\langle v,k,\lambda\rangle$ and compute $b$ and $r$ in terms of the former three. However, although these relations restrict the set of admissible parameters for a BIBD, such admissibility is a necessary yet insufficient condition to guarantee its existence \cite{cox,yates2}. 

The BIBD problem is a constraint satisfaction problem (CSP) that can be readily transformed into a constraint optimisation problem (COP) by relaxing the problem (allowing the violation of constraints) and defining an objective function that accounts for the number and degree of their violations. More precisely,  let  $I = \langle v,b,r,k,\lambda\rangle$ and $M$ represent, respectively, the instance values and the (binary) incidence matrix of size $v \times b$ for a BIBD problem. In addition, for the rest of the paper, let  $\mathbb{N}^{+}_{h}=\{1,...,h\}$ (for any integer number $h\ge 1$). Therefore, the problem of finding a BIBD solution can be formulated as follows:

	\begin{equation}
	\min \; f^I(M) = \sum_{i=1}^{v} \phi_{i}(M,r)+ \sum_{j=1}^{b} \phi'_{j}(M,k) + \sum_{i=1}^{v-1} \sum_{j=i+1}^{v} \phi''_{ij}(M,\lambda) 
	\label{primal_r constraints}
	\end{equation}
where 
	\begin{eqnarray}
	\label{eq1}
	\phi_{i}(M,r)=\left| r - \sum_{j=1}^{b}m_{ij}\right|, \qquad\qquad \forall i \in [1,v]\\
	\label{eq2}
	\phi_{j}'(M,k)=\left| k - \sum_{i=1}^{v}m_{ij}\right|, \qquad\qquad \forall j \in [1,b]\\
	\label{eq3}
	\phi_{ij}''(M,\lambda)= \left| \lambda - \sum_{h=1}^{b}  m_{ih}m_{jh}\right|, \quad\forall i,j \in [1,v]:i<j
	\end{eqnarray}

We call this formulation {\em the primal model}, denoted as $B$ because it is based on a binary representation of the candidates to be solved.  Note that the required values of the  row constraints, column constraints and scalar product constraints correspond with the number of ones per row (i.e. $r$), the number of ones per column (i.e. $k$), and the scalar product of any pair of different rows (i.e. $\lambda$). So, for each row $i$ (resp. column $j$) in the incidence matrix, $\phi_{i}(M,r)$ in Eq. (\ref{eq1}) (resp. $\phi_{j}'(M,k$) Eq.(\ref{eq2})) computes the discrepancies between the required value $r$ (resp. $k$) of ones for row $i$ (resp. column $j$) and the existing number of ones in row $i$ (resp. column $j$). Also note that for each pair of distinct rows $i,j$  in the incidence matrix, $\phi_{ij}''(M,\lambda)$ in Eq. (\ref{eq3})  calculates the discrepancies between the required value $\lambda$ of the scalar product of the rows (i.e. coincidences of ones placed in the same positions in both rows) and the computed scalar product  of the two rows (i.e. the existing number of ones in the same position in both rows $i$ and $j$). As a consequence, for a given incidence matrix $M$, the value returned by the objective function sums up all discrepancies with respect to
the required values of the row constraints (i.e. Eq.(\ref{eq1})), column constraints (i.e. Eq.(\ref{eq2})) and scalar product constraints (i.e. Eq.(\ref{eq3})).

Then, a solution to the BIBD problem is  a configuration $M^*$ such that $f^I(M^*)=0$.

\subsection{Related work}
\label{sect:related work}
The BIBD problem has been tackled by a number of different techniques in the literature, with varying levels of success. Traditionally, the problem has been dealt with deterministic, constructive and/or complete methods.  For instance, Whitaker \emph{et al.} \cite{WhitakerTriggs1990} used mathematical programming methods to look for an optimal incomplete block design. Zergaw \cite{zergaw} also considered the error correlation, and presented a sequential algorithm for constructing optimal block designs. Along the same lines, Tjur \cite{tjur} incorporated interchange mechanisms with the addition of experimental units (blocks) one by one. Flener \emph{et al.} \cite{flener01} proposed a matrix model based on ECLIPSE to solve the problem of block generation. 

One of the key points of interest in the problem is its symmetrical nature (i.e. rows and columns can be permuted and objects can be relabelled). In this sense, constraint programming is the  most frequently used technique to deal with this issue.  For instance, Puget \cite{DBLP:conf/cp/Puget02} formulated the problem as a CSP where each instance was represented by a classical binary matrix of size $v \times b$, and proposed combining methods for \emph{symmetry breaking via dominance detection} and \emph{symmetry breaking using stabilisers} in order to solve the problem. In addition, Meseguer and Torras~\cite{DBLP:journals/ai/MeseguerT01} explored two strategies (namely, a heuristic for variable selection and a domain pruning procedure) to exploit the symmetry of the problem. The underlying idea in this approach was to use symmetries to guide the search for a solution. The objective was not to solve specific instances but rather to be effective in reducing the search effort~\cite{colbourn2010crc}.

Although all these methods can be used to design BIBDs, their applicability is limited by the size of the problem instances. To address this, stochastic methods have also been applied to the problem. For instance, Bofill et al. \cite{DBLP:journals/nn/BofillGT03} formulated the generation of BIBD as a combinatorial optimisation problem tackled with a neural network. A simulated annealing algorithm endowed with this neural network (NN-SA) was shown to offer better performance than an analogous hybridisation with mean field annealing. These results were further improved upon by Prestwich \cite{DBLP:conf/cp2003/PrestwichCP03,prestwich:negative-efefcts-aor03},  who considered different schemes for adding symmetry breaking constraints inside a constrained local search (CLS). 

These results were improved by Rodr\'{\i}guez \emph{et al.} \cite{RuedaCF09} who used both local search methods (hill climbing and tabu search) and population-based techniques (genetic algorithms). Two different neighbourhood structures (defined over the primal encoding described in Sect. \ref{subsec:primal model}) were proposed, one based on bit-flipping and the other on position-swapping. It was shown that the swap-based neighbourhood (sw) was superior to the flip-based neighbourhood, and that tabu search based on position-swapping
(TS$_{\rm sw}$) offered the best performance, being capable not only of beating hill climbing methods --also based on the swap-based neighbourhood-- and genetic algorithms (GA$_{\rm sw}$) --based on both position-swapping and the bitwise uniform crossover operator ({UX})--  in more than 78\% of the instances, but also of solving 57 instances from a selected set of 86 ($66.28 \%$), one more than CLS, the best method until that moment. Later,  Rodr\'{\i}guez \emph{et al.} \cite{RuedaCF09} explored, in greater depth, the use of population-based methods \cite{DBLP:journals/ijcopi/RuedaCL11}, and more specifically, the use of memetic algorithms (MAs) in the form of a synergistic combination of a genetic algorithm with the use of a heuristic recombination operator and a TS-based local searcher. It was shown that MAs with a greedy recombination operator ({Gd}), termed  MA$_{\rm Gd}$, performed better than MAs based on the bitwise uniform crossover operator ({UX}) (as proposed in \cite{RuedaCF09}) as well as its constituent parts TS$_{\rm sw}$ and GA$_{\rm sw}$ (as described in \cite{RuedaCF09}). By increasing the number of evaluations, MA$_{\rm Gd}$ solved  63/86 instances ($73.26\%$), TS$_{\rm sw}$ 59/86, and GA$_{\rm Gd}$ (i.e. a GA with the operator Gd) 48/86. MA$_{\rm Gd}$ can be considered the state-of-the-art (non-commercial) heuristic method.

There are other approaches which address the generation of BIBDs. In particular, there have been several constructive approaches such as the one described by Yokoya and Yamada \cite{yokoya2011mathematical}. These authors  made use of the power of the commercial linear programming solver CPLEX and proposed a non-linear mixed-integer programming approach that was shown to be effective in solving  BIBD instances. Later, Mandal \cite{Mandal2014183} presented an improved linear integer programming approach that handled the problem in an easier way. Along the same lines, Rodr\'{\i}guez \emph{et al.} \cite{david:jorunal_colombian_2016} also proposed a constructive approach, based on local search with multi-start, that provided a greater capacity of exploration of new zones (major diversification) unlike, for example, other approaches that use major intensification. This method has shown good performance, although there were instances for which it was not able to reach optimal solutions. These methods, although efficient in finding  solutions, all demand a high computational effort to generate BIBDs. Moreover, they follow a constructive approach that is very different from our metaheuristic proposals. Consequently, they are not considered here in our experimental study. 

Table \ref{tab:PorcInsResueltas86} summarises the performance (measured in number of problem instances solved) of all the metaheuristics methods mentioned in this section. We do not provide running times as this information was not reported for all the methods and, in addition, it strongly depends on many external factors.
   
\begin{table}[!t]
\caption{Number (\#) and percentage (\%) of problem instances solved by the basic and integrative metaheuristics (identified in first column) working alone on the set of 86 instances taken from \cite{DBLP:conf/cp2003/PrestwichCP03,DBLP:journals/nn/BofillGT03}. The third column indicates the reference in which the method was reported. } 
\label{tab:PorcInsResueltas86}
       \begin{center}
           \begin{tabular}{|c|c|c|}
               \hline
               algorithm             & \# (\%)             & Ref.\\
               \hline
               NN-SA           &  16 ($18.60\%$) & \cite{DBLP:journals/nn/BofillGT03}\\
               CLS             & 56 ($65.12\%$)  & \cite{prestwich:negative-efefcts-aor03,DBLP:conf/cp2003/PrestwichCP03}\\
               TS$_{\rm sw}$         &  57 ($66.28\%$) & \cite{RuedaCF09} \\
               GA$_{\rm sw}$         &  37 ($43.02\%$) & \cite{RuedaCF09} \\
               TS$_{\rm sw}$         &  59 ($68.60\%$) & \cite{DBLP:journals/ijcopi/RuedaCL11} \\
               GA$_{\rm Gd}$         &  48 ($55.81\%$) & \cite{DBLP:journals/ijcopi/RuedaCL11} \\
               MA$_{\rm Gd}$   & {\bfseries{63}} ($73.26\%$)  &   \cite{DBLP:journals/ijcopi/RuedaCL11} \\
               \hline
               TABU-BIBD(20)   & 78 ($90.70\%$) &  \cite{yokoya2011mathematical} \\
               Multi-step  & {\bfseries{79}} ($91.86\%$) & \cite{Mandal2014183} \\
               Ts+Hc  & 74 ($86.05\%$) & \cite{david:jorunal_colombian_2016}  \\
               \hline
           \end{tabular}
       \end{center}
   \end{table}

\section{Solving the BIBD with metaheuristics}
\label{sect:metaheuristics}
With the aim of keeping this paper relatively self-contained, this section describes part of our previous work on the application of metaheuristics to the BIBD problem; in particular we outline the work described in \cite{RuedaCF09,DBLP:journals/ijcopi/RuedaCL11}, where the primal problem representation (i.e. the binary encoding) was used.

In \cite{RuedaCF09}, two neighbourhood structures were considered: the first arose naturally from the binary representation of solutions using the incidence matrix $M$ and was  based on
the Hamming distance (\textsf{bit-flip (bf)}); the second (denoted as \textsf{swap (sw)}) took an object from one block, and moved it to a different one, which can be formulated in binary terms as permuting a 0 and a 1 within the same row. Then, three different techniques based on these two neighbourhood variants, were proposed; more specifically, two hill climbing (HC) methods (i.e. steepest descent procedures termed HC$_{bf}$ and HC$_{sw}$), two tabu search algorithms (TS$_{bf}$ and TS$_{sw}$), and two steady-state genetic algorithms based on  binary tournament selection and replacement of the worst individual in the population (termed GA$_{bf}$ and GA$_{sw}$); GA$_{bf}$ used uniform crossover and bit-flip mutation --easy to implement in the binary space, whereas GA$_{sw}$ used uniform crossover at row level (that is, it randomly selects entire rows from either parent) and swap mutation (the interested reader is referred to \cite{RuedaCF09} for more details). For the experiments, 86 instances taken from \cite{DBLP:conf/cp2003/PrestwichCP03} were used as benchmarks. The best proposal was TS$_{sw}$ which solved 57 instances from those 86, and performed better than CLS \cite{DBLP:conf/cp2003/PrestwichCP03}, the previous best solution method (see Table \ref{tab:PorcInsResueltas86}).

\label{subsect:MAs}
Subsequently, in \cite{DBLP:journals/ijcopi/RuedaCL11} we proposed a memetic algorithm (MA) \cite{HandbookMemeticNeriCotta2012} to finding BIBDs.  The resulting MA could be characterised as a steady-state genetic algorithm (GA) --which serves as the underlying population-based search mechanism-- based on the $sw$ scheme defined before (hence, intrinsically enforcing the row constraint), that incorporates two intensifying components, namely a specific heuristic recombination operator, and a local searcher, in order to guide the search towards promising regions in the search space. More precisely, 
two different multi-parent recombination operators, one based on uniform crossover and termed UX, and a specific greedy version of the uniform crossover termed Gd, were proposed. The {Gd} operator starts by creating a set with all available rows in the parents; then (if there are enough different rows; otherwise, standard {UX} is invoked), it randomly selects an initial row, and subsequently tests all available rows, picking the one which violates fewer scalar-product constraints. It was shown that the recombination operator played an important role in the discovery of new improved solutions. The local search (LS) method used inside the MA to intensify the search  was TS$_{sw}$ (mentioned above). More details can be found in \cite{DBLP:journals/ijcopi/RuedaCL11}. 

The general scheme of the MA is depicted in Algorithm~\ref{fig:MA}. The input of this algorithm consists of the problem parameters (i.e. $v$, $b$, $r$, $k$, and $\lambda$), as well as the algorithm parameters, i.e the genetic operators and their associated parameters (such as, e.g. application rates); the output of the algorithm is the best individual found during the search process. Parent selection was done randomly using a binary tournament for breeding and replacement of the worst individual in the population. To preserve the diversity in the population, no duplicate solution was accepted, and a re-starting mechanism re-activated the search whenever stagnation occurred. This was done by keeping a fraction $f\%$ of the top individuals in the current population,  and refreshing the rest of the population with random individuals (line 19). This procedure was triggered after a number of evaluations without any improvement in the current best solution ($n_\iota$). The local search was restricted to explore $n_{\nu}$ neighbours, and $p_{LS}$, $p_X$ and $p_M$ represent the probability of applying local search, the crossover and mutation rates, respectively.

\begin{algorithm}[!ht]
		\caption{Pseudo-code of the memetic algorithm.}
		\label{fig:MA}
		\Begin {
			\For {$i \leftarrow 1$ \textbf{to} $popsize$}
			{
				$pop[i]$ $\leftarrow$ \textsc{GenerateMatrix}($v, b, r$)\;
				\textsc{Evaluate}($pop[i]$)\;
			}
			\While {$numEvals < maxEvals$}
			{
				\eIf {rand $<p_{X}$}
				{
					$parent_1$ $\leftarrow$ \textsc{TournamentSelect}($pop$)\;
					\textsc{...}\;
					$parent_m$ $\leftarrow$ \textsc{TournamentSelect}($pop$)\;
					$\mathit{offspring}$ $\leftarrow$ \textsc{Recombine}($parent_1$,$\ldots$,$parent_m$)\;
				}
				{
					$\mathit{offspring}$ $\leftarrow$ \textsc{TournamentSelect}($pop$)\;
				}
				$\mathit{offspring}$ $\leftarrow$ \textsc{Mutate}($\mathit{offspring}, p_M$)\;
				\lIf {rand $<p_{LS}$}
				{
					$\mathit{offspring}$ $\leftarrow$ \textsc{LocalSearch}($\mathit{offspring}, n_{\nu}$)
				}
				\textsc{Evaluate}($\mathit{offspring}$)\;
				$pop$ $\leftarrow$ \textsc{Replace}($pop$, $\mathit{offspring}$)\;
				\lIf {stagnation($n_\iota$)} {
					$pop$ $\leftarrow$ \textsc{Restart}($pop$,$f\%$)
				}
				
			}
		}
	\end{algorithm}

It was shown that algorithms with the {Gd} operator solve more instances than their counterparts with {UX}, and MAs also outperform their GA counterparts. The version MA$_{\rm Gd}$ solved 63/86 instances, which even better than TS$_{sw}$, Even though a higher number of evaluations (i.e. $2 \times 10^7$) was considered in \cite{DBLP:journals/ijcopi/RuedaCL11} than the one given in \cite{RuedaCF09} (i.e, $2 \times 10^6$), MA$_{\rm Gd}$ can be considered the best metaheuristic solution reported so far in the literature.

\section{A new battery of metaheuristics based on symmetry breaking, dual models and hybridisation}
This section presents a number of new metaheuristic proposals to handle the BIBD problem. In Sect.~\ref{sec:DualModel}, we first propose a novel dual problem representation for the BIBD, and a new problem formulation based on it. Next, Sect.~\ref{sec:symetriesbreaking} describes a symmetry-breaking method (that, to the best of our knowledge, is also novel for the problem under consideration in this paper) for both the primal representation (i.e. the classical binary one) and its dual encoding (i.e. the decimal representation). Finally, in Sect.~\ref{Sec:EsquemasCooperativos}, we suggest a cooperative scheme as an alternative to the integrative memetic algorithm (Algorithm \ref{fig:MA}). One of the primary particularities of this cooperative scheme is that it allows the cooperation of algorithms designed to work  on different representation models (i.e. primal or dual).

\subsection{A dual representation}
\label{sec:DualModel}
\begin{figure}[t]
	\begin{center}
		\includegraphics[width=0.7\columnwidth]{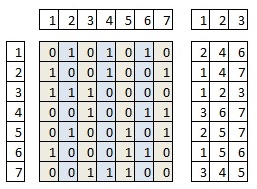} 
	\end{center}
	\caption{Primal/Binary and dual/decimal encodings of the $\langle 7,7,3,3,1\rangle-$symmetric BIBD shown in Figure~\ref{fig:instanciaBIBD} (b). The number of columns (7 and 3 for the primal and dual representations, respectively) and rows (7 in both cases) are identified for clarity.} 
	\label{fig:dual representation: example}	
\end{figure}

It is well-known that the representation of candidate solutions can have dramatic effects on the problem solving process, especially in the universe of evolutionary algorithms \cite{DBLP:books/daglib/0014740}. For this reason, we consider the concept of {\em duality} as a way to obtain alternative representations to the natural (and primal) encoding of the solutions to the BIBD problem. Hence, the alternative model we propose is termed {\em the dual model} or {\em decimal formulation (D)};
Figure \ref{fig:dual representation: example} shows an example of a dual representation for a given symmetric BIBD instance.
Basically,  the solution (or candidate) to the BIBD is  now defined by a dual incidence matrix $M^d \equiv \{m^d_{ij}\}_{v \times r}$, which is a $v \times r$ integer matrix where $m^d_{ij} \in \mathbb{N}^{+}_{b}$ contains a value from the range $[1,b]$ which identifies a block containing the object $i$. Note that there are $r$ columns (i.e. $j \in [1,r]$), so that  each object $i$ is contained exactly in $r$ blocks 
if a constraint that all values in a row have to be different is imposed.  
The dual formulation of the BIBD problem corresponds to a relaxed CSP problem with an objective function that involves the number and degree of violations of constraints defined only on the parameters $k$ and $\lambda$ as follows:

	\begin{equation}
	\label{dual function to minimize}
	\min \; f^I_d(M^d) = \sum_{j=1}^{b}\psi_j(M^d,k) + \sum_{i=1}^{v-1} \sum_{j=i+1}^{v} \psi'_{ij}(M^d,\lambda) 
	\end{equation}
such that every object should be assigned to $r$ distinct blocks, that is to say:
	\begin{equation}
	\label{dual rep:all-different constraints}
	\forall j,h\in \mathbb{N}^{+}_{r}:  j \neq h \Rightarrow m^d_{ij}\neq m^d_{ih}, \qquad \forall i\in [1,b]
	\end{equation}
where
	\begin{eqnarray}	
	\label{dual model: k discrepancies}
	\psi_{j}(M^d,k)=\left| k-\sum_{i=1}^{v}\sum_{h=1}^{r} [m^d_{ih}=j] \right|, \qquad\forall j \in [1,b]\\ 
	\nonumber \\ 
	\label{dual model: lambda discrepancies}
	\psi_{ij}'(M^d,\lambda)= \left| \lambda - \sum_{h=1}^{r} \sum_{l=1}^{r} [m^d_{ih}=m^d_{jl}]\right|, \qquad\qquad 
    \\ \nonumber
	\qquad\qquad\forall i,j \in [1,v]: i < j 
	\end{eqnarray}

In Eq. (\ref{dual model: k discrepancies})  and Eq. (\ref{dual model: lambda discrepancies}) we employ the Iverson brackets [] (i.e. [P]=1 if $P$ is true, and 0 otherwise). Observe that, by adding an all-different-value constraint associated with each row -- i.e. constraint (\ref{dual rep:all-different constraints})-- each row $i$  now contains the $r$ (required) block assignments of object $i$, and thus the constraint requiring that each object be placed in $r$ blocks is implicitly present in the new dual model. Therefore, the function to be minimised -- i.e. (\ref{dual function to minimize}) -- only has two components. The first one, shown in Eq. (\ref{dual model: k discrepancies}),  sums the discrepancies from the value $k$. Note that $\sum_{i=1}^{v}\sum_{h=1}^{r} [m^d_{ih}=j]$ quantifies how many times the $j$th block appears in the solution and each block $j \in [1,b]$ has to contain exactly $k$ objects, something that happens when $\psi_{j}(M^d,k)$ equals 0. The second component of  the objective function, shown in Eq. (\ref{dual model: lambda discrepancies}), computes the discrepancies from the value $\lambda$. Observe that each object $i$ has to coincide with any other object $j$ in exactly $\lambda$ blocks, which means that each two rows of $M^d$ have to share $\lambda$ blocks (as in the primal model). Hence, we measure the discrepancies between any two objects in (\ref{dual model: lambda discrepancies}) with respect to the required value $\lambda$.
A solution to the BIBD problem is thus a configuration $M^{d*}$ such that $f_d^I(M^{d*})=0$. 

In this model the neighbourhood is similar to the ${swap}$ version considered for the primal scheme as defined in \cite{RuedaCF09}, that is to say, a neighbour of a matrix $M^d$ is any other incidence matrix $M'^d$ obtained from $M^d$ by replacing an element $\phi \in \mathbb{N}^{+}_{b}$ contained in a cell $m^d_{ij}$  by any other label in $\mathbb{N}^{+}_{b}\setminus\{\phi\}$, provided constraint 
(\ref{dual rep:all-different constraints}) is still satisfied. 

\subsection{Symmetry breaking}
\label{sec:symetriesbreaking}
According to \cite{fahle2001}, one way of reducing a problem's symmetries is to transform it into another problem with the same characteristics as the original  but eliminating all or most symmetrical states.  In the last few decades, a number of methods have been applied to deal with the problem of symmetry \cite{benhamou1994study,fahle2001,Gent99symmetrybreaking,DBLP:journals/ai/MeseguerT01}. Most of these methods are primarily aimed to reduce the search space of the problem. Other recent works have shown how solving combinatorial problems via mixed integer linear programming approaches can be sped up by adding symmetry breaking constraints to the original formulation. Another idea is to consider asymmetric representatives formulations (ARF) as alternatives to the natural symmetric formulation of the problem. They have been shown to be effective to deal with combinatorial optimisation problems such as p job grouping, binary clustering, node colouring, or blocking experimental designs \cite{CAMPELO20081097,G201044,JANS20131132,G2016117}. 

In this paper, we consider a symmetry-breaking approach, both for the primal model and for the dual model. This approach is called variable reduction in \cite{G2016117}. 

\subsubsection{Primal (or binary) model}
\label{sec:SymmetryBreakingPrimalModel}
Consider the problem representation introduced in Sect.~\ref{subsec:primal model}. In general, BIBD symmetries arise, first of all, because any two objects are interchangeable in the sense that any two rows can be permuted (i.e. the corresponding objects can have their labels swapped) in the incidence matrix and the resulting candidate will be the same solution. For primal encoding, in particular, this argument can be extended  to blocks/columns,  as any two columns can be permuted as well. To tackle these symmetries, we impose the following four constraints on the primal problem formulation:

\begin{itemize}
	\item In Row 1: set $m_{1j}=1$ for each $j \in \mathbb{N}^{+}_{r}$ and $m_{1j}=0$ for $r < j \leq b$. In other words, place the first object (row 1) in the first $r$ blocks (i.e. the first $r$ columns) so that the row constraint is satisfied for object 1.
	
	\item In Row 2: set $m_{2j}=1$ for $j \in \mathbb{N}^{+}_{\lambda}$, $m_{2(\lambda+j')}=0$ and $m_{2(r+j')}=1$ for $j' \in \mathbb{N}^{+}_{r-\lambda}$, and 
	set the other cells in row 2 to 0. 
	This
	guarantees that the scalar product constraint between rows 1 and 2 is satisfied.
	
	\item In Column 1: set $m_{i1}=1$ for each $i \in \mathbb{N}^{+}_{k}$ and $m_{1i}=0$ for $r < i \leq v$. In other words, place the first $k$ objects in the first block (i.e. the first column) so that the column constraint is satisfied for the first block.
	
	\item In Column 2: set $m_{i2}=1$ for $v-k-(m_{12}+m_{22})< i \leq v$, and the other values in the column (except the first two rows, to 0). In other words, taking into account that $(m_{12}+m_{22})$ objects have already been placed in block 2, we place the last $k-(m_{12}+m_{22})$ objects in block 2, so that the column constraint is satisfied.
\end{itemize}

\begin{figure}[t]
	\begin{center}
		\includegraphics[width=1.0\columnwidth]{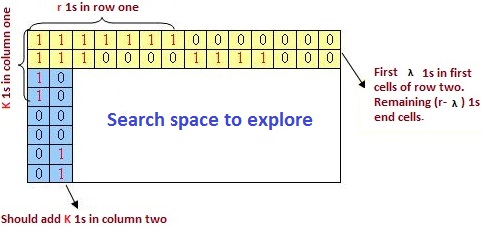}
	\end{center}
	\caption{Symmetry breaking in an $\langle 8,14,7,4,3 \rangle$-instance (binary encoding.)	\label{fig:SB}
}
\end{figure}

These four constraints can also be viewed as a preset process that fixes the values of the first two rows and the first two columns in the incidence binary matrix $M \equiv \{m_{ij}\}_{v \times b}$ of a particular $\langle v,b,r,k,\lambda\rangle$-instance. This produces a slight reduction of the problem symmetries. Note that the first two rows and the first two columns remain constant in each candidate solution so that they will never be permuted with any other row (resp. column). As a direct consequence,  the search space is also reduced. Fixing these two rows/columns means that optimisation only has to be conducted in a binary matrix of size ${(v-2) \times (b-2)}$. Figure~\ref{fig:SB} shows an example of how to fix the rows and columns for an $\langle 8,14,7,4,3 \rangle$-instance in the binary problem representation.

\subsubsection{Dual (or decimal) model}
\label{sec:BreakingSDualModel}
Now, consider the dual problem representation introduced in Sect.~\ref{sec:DualModel}. Symmetry breaking is achieved by fixing the two first rows of the incident matrix $M^d$ as follows: 

\begin{figure}[t]
	\begin{center}
		\includegraphics[width=0.7\columnwidth]{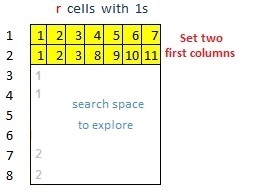}
	\end{center}
	\caption{Symmetry breaking in the dual model of an 
	$\langle 8,14,7,4,3 \rangle$-instance. The first two rows correspond to the dual representation of the first two rows in the primal representation shown in Figure~\ref{fig:SB}.}
	\label{fig:SBDual}
\end{figure}

\begin{itemize}
	\item In Row 1: set $m_{1j}=j$ for $j \in \mathbb{N}^{+}_{r}$. In other words, the first row contains numbers from 1 to $r$ in sequence. This constraint assures that the first object (i.e. row 1) is placed in the first $r$  blocks and breaks the symmetry of object placement.	
	\item In Row 2: set $m_{2j}=j$ for $j \in \mathbb{N}^{+}_{\lambda}$, and set $m_{2j'}=r+j'-\lambda$ for $j'>\lambda$.	
	In this way, the second object is placed in the first $\lambda$ blocks (where the first object is also placed), and also in $r-\lambda$ blocks other than those in which the first object was placed. This constraint guarantees that the scalar product constraint is satisfied for objects 1 and 2.
\end{itemize}

The idea is to fix the values of the first two rows in the incidence dual matrix $M^d$ (of size $v \times r$) of a particular $\langle v,b,r,k,\lambda\rangle$-instance so that optimisation only has to be conducted in an integer matrix of size ${(v-2) \times r}$. Note that, unlike in the primal model, we have not fixed the columns in the dual formulation. The reason is that the first column would only be partially completed anyway, since, in the dual formulation, it is not possible to specify that an object (e.g. objects 5 and 6) is not placed in some specific block (e.g. blocks 1 and 2). This is possible in the primal model by using the value 0 in a cell of the matrix (see Figure~\ref{fig:SB}).
In addition, from an implementation point of view, partially completing the first columns would also hinder the natural encoding of individuals as a rectangular matrix. Figure~\ref{fig:SBDual} shows an example of how to fix the two first rows in the dual encoding of a problem instance $\langle 8,14,7,4,3\rangle$.

\subsection{Cooperative model architecture}
\label{Sec:EsquemasCooperativos}
The memetic algorithm described in Sect. \ref{sect:metaheuristics}, and presented in Algorithm~\ref{fig:MA}, can be viewed, according to the taxonomy for hybrid and cooperative algorithms given by Puchinger and Raidl \cite{Puchinger2005}, as an integrative hybrid algorithm in which a local search is subordinated to the execution of an external genetic algorithm (GA). In other words, local search is executed inside a GA. Puchinger and Raidl presented another interesting scheme, the {\em collaborative approach} in which several optimisation algorithms are executed in parallel (or sequentially) and exchange information with certain frequency.	
This kind of cooperation can be considered in itself a programming paradigm comprising two main elements \cite{Crainic2008}: (a) a set of autonomous programs (usually called agents), each implementing a particular solution method, and (b) a cooperative scheme that combines these autonomous elements into a simple and unified strategy for troubleshooting. In 
this collaborative approach, 
the idea is to apply a number of (possibly different) optimisation algorithms each of which explores a  specific part of the search landscape through processes of intensification.
Next, the agents synchronise from time to time to exchange information.  A specific spatial structure  (e.g. a ring in which each  agent has a predecessor and a successor) identifies the communication topology, that is to say, the way in which this information is transmitted between the agents (e.g. the information is transmitted from any given agent to its successor in the ring-based structure). 
The set of agents involved in this collaborative scheme can be regarded as a network of nodes, each  containing a certain optimisation algorithm. These algorithms (i.e. the agents) operate in different parts of the same search space. The whole scheme constitutes an effective mechanism for escaping from local minima (by means of the information exchange among the agents).
This approach has been proven  to be efficient for a number of combinatorial problems \cite{Cruz2009,Masegosa2009,Amaya2011,Amaya2011MC}.

\begin{algorithm}[!t]
		\caption{{\sc Cooperative-Model$_n$}}\label{alg:model}
		\label{alg:cooperative algorithm}
		\For{$i\in\mathbb{N}^+_{n}$} {
		\tcp{Generation Adjusted to the problem model tackled by agent $a_i$}
			$S_i \leftarrow$ \textsc{GenerateInitialPopulation()};  
		}
		$cycles \leftarrow 1$\;
		\While{$cycles \leq \Theta$} {
			\For{$i\in\mathbb{N}^+_{n}$} {
			     \tcp{Population update}
				$S_i \leftarrow {a}_i(S_i)$;
			}
			\For{$(i,j)\in \mathbf{T}_R$} {
				\tcp{Select candidate to migrate via the migration policy}
				$s_{submitted} \leftarrow \textsc{selectCandidateFrom}(S_i)$ \; 
				\tcp{Now, test candidate acceptance via the acceptance policy}
				\If{$\textsc{AcceptSubmittedCandidate}(s_{submitted}, S_j)$}
				{
				    \tcp{Selection of candidate to replace}
					$s_{tobereplaced} \leftarrow	\textsc{selectCandidateToReplaceIn}(S_j)$\; 			\tcp{adding migrated candidate (translated to the problem encoding of agent $j$)}	
					$S_j \leftarrow S_j \cup \{$\textsc{Encoding}$_j(s_{submitted})\} \setminus$ \{$s_{tobereplaced}$\} \;
				}
			}
			$cycles \leftarrow cycles + 1$\;
		}
		\Return{$\arg\min\{\textsc{Fitness}($\textsc{Best}$(S_i)) \ |\ i \in \mathbb{N}^+_n\}$}\;
\end{algorithm}

Now, unlike what it is often done in this kind of collaborative approach, we study the effect of considering different search spaces to be handled separately in the nodes  of the network. More precisely, we consider a number of cooperative algorithms in which agents  are loaded with one of the techniques  previously proposed in Sect.~\ref{sect:metaheuristics} or their equivalent adapted to the dual representation (as shown in Sect.~\ref{sec:DualModel}), and where the method loaded in any given agent might also have symmetry breaking constraints (as explained in Sect.~\ref{sec:symetriesbreaking}).   This means that some agents possibly work on different encoding/search spaces and use distinct problem formulations. The algorithms depend on their interaction topology and the model used for encoding the candidates, and these are discussed below.

\subsubsection{Formal definition}

Let $R$ be an architecture with $n$ agents; each agent $a_i$ ($1 \leq i \leq n$) in $R$ consists of one of the metaheuristics described in preceding sections. Therefore, these agents can work on the primal or dual model, with or without symmetry breaking. The agents engage in periods of isolated exploration followed by synchronous communication. We denote by $\Theta$ the maximum number of exploration/communication cycles in a certain cooperative model. In addition, let $S_i$ be the pool of solution candidates associated with agent $a_i$ (i.e. if the agent is loaded with a local search (LS) method then $\#S_i = 1$, and if the agent is endowed with a population based method --e.g. an MA-- then $\#S_i \geq 1$, where $\#S_i$ represents the cardinality of $S_i$), and let $\mathbf{T}_R\subseteq\mathbb{N}^+_n\times\mathbb{N}^+_n$ be the communication topology over $R$ (i.e. if $(i,j)\in\mathbf{T}_R$ then agent $a_i$ can send information to agent $a_j$). The general architecture of the model is described in Algorithm \ref{alg:model}.  This algorithm's input consists of the problem parameters (i.e. $v$, $b$, $r$, $k$, and $\lambda$), the topology of the agent network (that defines the communication policy, as explained below),  the $n$ algorithms (i.e. agents or metaheuristics) running on each node of the cooperative system (i.e. the network), the candidate migration policy, and the  criteria for accepting the candidates  (see below for  details about these two procedures). Each  agent also has its own parameters (such as operator application rates, and type of encoding --primal/dual--). The algorithm's output is the best individual found during the search process (line 18; note that the  $\textsc{Fitness}$ function is a well-known concept in evolutionary computing and basically returns a value that measures how close a candidate is to an optimal solution). First, all the agents are initialised with random solution(s) (lines 1-3). The initialisation of a pool $S_i$  associated with agent $i$ in the system is specific to the model (i.e. primal or dual) handled by the agent $a_i$. Next, the algorithm is executed for a maximum number $\Theta$ of iteration cycles (lines 5-15) where, in each cycle, the search technique contained within each agent is executed to update its associated pool of solutions (lines 6-8);
note also that, if the agent contains an LS method, this basically means an improvement of its unique solution, but if the agent contains a population-based method, then a new pool of solutions is generated).
Next, solutions are fed from one agent to another according to the topology considered (lines 9-14). This process means that, initially (line 10), the candidate to be transmitted from the pool of the source agent (i.e. node or metaheuristic $i$) is selected with respect to the {\em migration policy} given as input (see below). Next, agent $a_j$ checks (line 11) whether the incoming solution from agent $a_i$ (line 10) has to be accepted according to the {\em acceptance criteria} also provided as input (see below).  Finally, if the submitted solution is accepted, it replaces an individual in the candidate pool of agent $j$ (line 13); the candidate to replace is selected via previously defined heuristics (line 12). Note  that many different  criteria for candidate migration (from  agent $i$ to agent $j$) and candidate acceptance (in agent $j$) can be defined. Combining diverse policies generates  different cooperative algorithms. We now describe a number of combinations that will be used in the experimental section.

\subsubsection{Communication topologies}

Three strategies for $\mathbf{T}_R$ (see line 9 in Algorithm~\ref{alg:model}) are considered here in this paper. These are based on the following interaction topologies:

\begin{itemize}
	\item \textsc{Ring}:
	$\mathbf{T}_R = \{(i, i(n) + 1) \mid i\in\mathbb{N}^+_n\text{ and
	}i(n)\text{ denotes }i\text{ modulo }n\}$. Thus, there exists a
	circular list of agents in which each agent only sends (resp.
	receives) information to its successor (resp. from its predecessor).
	
	\item \textsc{Broadcast}: $\mathbf{T}_R=\mathbb{N}^+_n\times\mathbb{N}^+_n$,
	i.e. a {\em go with the winners}-like topology in which the best
	overall solution at each synchronisation point is transmitted to all
	agents. This means all agents execute the intensification over the same part of the search space at the beginning of each cycle.
	
	\item \textsc{Random}:
	$\mathbf{T}_R$ is composed of $n$ pairs $(i,j)$ that are randomly
	sampled from $\mathbb{N}^+_n\times\mathbb{N}^+_n$. This sampling is
	done each time communication takes place, and, hence, any two agents
	might eventually communicate at any step.
\end{itemize}

These communication topologies were already proposed in  \cite{Amaya2011,Amaya2011MC} to handle a tool switching problem with some success; however, symmetry breaking, different encodings  and different policies were not considered for migration and solution acceptance in that work. Now, we propose  a wider cooperative scheme to handle symmetrical constrained optimisation problems. Moreover, we have also adapted some of the ideas proposed in \cite{NoguerasCotta2014} for memetic algorithms to our cooperative algorithms generated from the scheme in Algorithm \ref{alg:cooperative algorithm}.  In particular,  we consider a number of  policies for the submission of  candidates from agent $i$ (i.e. the migration policy) as well as the acceptance of candidates submitted to agent $j$ (i.e. the reception/acceptance policy). With respect to {\em candidate selection in the migration procedure} 
(i.e. $\textsc{selectCandidateFrom}(S_i)$ in line 10, of Algorithm \ref{alg:cooperative algorithm}),
we propose three  strategies:

\begin{itemize}
	\item (\textsc{Random} R): send a random solution of the pool from agent $i$, 
	\item (\textsc{Diverse} D): send the candidate in $S_i$ that maximises the diversity\footnote{\label{pie1}To this end, individuals whose genotypic distance (in a Hamming sense) to individuals in the receiving population is maximal are selected.} in $S_j$, and 
	\item (\textsc{Worst} W): send the worst candidate of the pool in agent~$i$. 
\end{itemize}

As for the reception and replacement policies (i.e. procedures $\textsc{AcceptSubmittedCandidate}(s_{submitted}, S_j)$ in line 11 and $\textsc{selectCandidateToReplaceIn}(S_j)$ in line 12, respectively), three alternatives are also considered: 
\begin{itemize}
	\item (\textsc{Random} R): always accept the submitted candidate and replace one random individual in pool $S_j$,
	\item (\textsc{Diverse} D): accept a new individual if and only if, it improves the diversity of the pool in agent $j$ and replace the worst, and 
	\item (\textsc{Worst} W): always accept the candidate and replace the worst in pool $S_j$. 
\end{itemize}

Also note that if a candidate solution taken from agent $i$ is finally accepted in agent $j$, it first  has to be translated -- if necessary -- to the encoding model used in agent $j$, as agents $a_i$ and $a_j$ may work on different search spaces (i.e. representation models); this is reflected in line 13 by the function called $\textsc{Encoding}_j(s_{submitted})$.

\section{Experiments}
\label{sec:experiments} 
This section describes the experimental analysis conducted. Given the large number of algorithms considered (resulting from the combination of different metaheuristics, encodings, use of symmetry-breaking procedures, communication topology, etc.), we first describe the notation in detail, as well as the experimental setting in Sect. \ref{sec:notation} and Sect. \ref{sec:configuration}, respectively. Subsequently, we report the results obtained in Sect. \ref{subsect:non-coop algo experiments} and Sect. \ref{subsect:coop algo experiments}, and analyse these in Sect. \ref{sec:analysis}.

\subsection{Notation}
\label{sec:notation}

In this subsection we explain the notation used to describe the algorithmic models, providing some specific examples for the sake of clarity.

\subsubsection{Non-cooperative algorithms}

Each algorithm is identified by a sequence of identifiers separated by a dot. First, the basic metaheuristics (as described in Sect. \ref{sect:metaheuristics}) are hill climbing (Hc), tabu search (Ts), genetic algorithm (GA), and memetic algorithm (MA), all of which are based on the swap neighbourhood. Additionally, for the population-based methods (i.e. GA and MA), the recombination procedure is characterised by the particular operator used --here we focus on the use of the greedy crossover operator (Gd)-- and by its arity, i.e. the number $m$ of parents used (denoted as A$m$). Additionally, we use an asterisk ($*$) to indicate the use of symmetry-breaking methods, a B to indicate that individuals are encoded in a binary way (i.e. the  primal model), and a D to indicate that these are encoded in the decimal representation (i.e.  the dual formulation). 

{\em Examples of notation of non-cooperative algorithms}: Hc.B (resp. Hc.B*) denotes a hill climbing method that was implemented for the primal (i.e. binary) model without (resp. with) symmetry breaking;  Ts.D (resp. Ts.D*)  denotes a tabu search implemented for the dual model without (resp. with) symmetry breaking; likewise, GA.B*.A2.Gd denotes a genetic algorithm with 2-parent greedy crossover (as explained in Sect. \ref{subsect:MAs} -- see \cite{DBLP:journals/ijcopi/RuedaCL11} for details on this crossover operator) implemented for the primal encoding with symmetry breaking, 
GA.D.A4.Gd is a genetic algorithm with a 4-parent greedy crossover implemented for the dual formulation without symmetry breaking, MA.Hc.B.A2.Gd a memetic algorithm with a hill climbing method as local search  and a 2-parent greedy  recombination implemented for the primal formulation without symmetry breaking, and MA.Ts.D*.A2.Gd is a memetic algorithm with tabu search as local search  and a 2-parent greedy crossover operator implemented for the dual model with symmetry breaking.

\subsubsection{Cooperative methods}

These algorithms are composed of some of the previous techniques combined according to given topology and migration policies. The notation $\mathbf{T}n(a_1,\ldots,a_n)$MR is used to characterise the method. Here:

\begin{itemize}
\item $\mathbf{T} \in \{\textsc{Broadcast}$ (Bc), $\textsc{Random}$ (Ra), $\textsc{Ring}$ (Ri)$\}$ denotes the topology of the model, 
\item ${n}$ is the number of agents (i.e. algorithms) connected as described in Sect. \ref{Sec:EsquemasCooperativos}, 
\item ${a}_i$ is the optimisation method used by agent $i$ (for $1 \leq i \leq n$), and
\item M, R $\in \{\textsc{Random}$ (R), $\textsc{Diverse}$ (D), $\textsc{Worst}$ (W) $\}$ identify,  respectively, the policies to migrate and accept candidates in the agents (see Sect. \ref{Sec:EsquemasCooperativos}). In our experiments, we have considered the following six combinations for $migration-reception$ policies: 
\begin{enumerate} 
\item \textsc{Random}-\textsc{Random} (RR), that is to say, \textsc{Random} policy for both migration and reception.
\item \textsc{Random}-\textsc{Worst} (RW): \textsc{Random} policy for migration and \textsc{Worst} strategy for reception.
\item \textsc{Random}-\textsc{Diverse} (RD): \textsc{Random} policy for migration and \textsc{Diverse} strategy for reception.
\item \textsc{Diverse}-\textsc{Random} (DR): \textsc{Diverse} policy for migration and \textsc{Random} strategy for reception. 
\item \textsc{Diverse}-\textsc{Worst} (DW): \textsc{Diverse} policy for migration and \textsc{Worst} strategy for reception, and 
\item \textsc{Diverse}-\textsc{Diverse} (DD): \textsc{Diverse} policy for migration and \textsc{Diverse} strategy for reception.
\end{enumerate}
Note that we do not include the combinations WD (i.e. \textsc{Worst}-\textsc{Diverse}), WR (i.e. \textsc{Worst}-\textsc{Random}) and WW (i.e. \textsc{Worst}-\textsc{Worst}). The reason is that preliminary experiments showed that choosing the \textsc{Worst} policy for migration exhibited a poor performance compared to the other combinations. 
\end{itemize}

{\em Examples of notation of cooperative algorithms}: Ri2(Ts.B, MA.Ts.D.A2.Gd)RW is a 2-agent \{\textsc{ring} topology\}-based cooperative algorithm that connects (a) a TS working on the binary representation, and (b) an MA that works on the dual representation, which uses a 2-parent greedy crossover, and that integrates TS as the underlying local search; in this case,  the algorithm always sends a random candidate selected from  the pool of the origin node (\textsc{Random} policy for migration), which will replace the worst individual in the destination node (\textsc{Worst} policy for the acceptance policy). Similarly, Ra3(Ts.B, MA.Ts.B.A2.Gd, MA.Ts.D.A4.Gd)RD denotes a 3-agent cooperative algorithm that connects, in a \textsc{random} topology, (a) tabu search and (b) two different MAs; the individuals to migrate are randomly chosen (i.e. a \textsc{Random} policy for migration) whereas candidates are accepted only if they increase the diversity of the solution pool (i.e. \textsc{Diverse} acceptance criteria). 

Note that in the cooperative algorithms the same optimisation method might be used by several agents (this is the case, for instance, in the algorithm Bc4(Ts.B,Ts.B,Ts.B,MA.Ts.D*.A4.Gd)RD in which 3 of the 4 agents contain the local search Ts.B.). The rationale for this is to try to increase the contribution of a certain method to the resulting cooperative hybrid, whose overall search profile is influenced by the particular mix of optimisation methods used. 
For clarity, in these cases,
we use the notation $\mathbf{T}n(pa, qb)$MR to denote the $n$-agent cooperative algorithm 
\[\mathbf{T}n(\underbrace{a,\ldots,a}_{p\ {\rm times}}, \underbrace{b \ldots b}_{q\ {\rm times}}){\rm MR}\] 
in which agents $a$ and $b$ are employed $p$ and $q$ times respectively (and where $p$ and $q$ are arbitrary numbers  that fulfill $n=p+q$);  moreover, $p$ (resp. $q$) is not written when $p=1$ (resp. $q=1$). So, for instance, Bc4(3Ts.B, MA.Ts.D*.A4.Gd)RD denotes the model Bc4(Ts.B,Ts.B,Ts.B,MA.Ts.D*.A4.Gd)RD  (i.e. here $p=3$ and $q=1$). Also,
Ra5(3Ts.B,2MA.Ts.B.A2.Gd)DW is a 5-agent algorithm where the local search Ts.B is embedded in 3 agents and the algorithm MA.Ts.B.A2.Gd is contained within the other two agents (i.e. here $p=3$ and $q=2$).

\subsection{Experimental configuration} 
\label{sec:configuration}
The experiments were conducted on the 86 instances taken from \cite{DBLP:conf/cp2003/PrestwichCP03,DBLP:journals/nn/BofillGT03} where $ vb \leqslant 1000$ and $k \neq 3$. This corresponds to the
hardest instances reported, since the cases where $k=3$ were easily solvable.  All algorithms have been run 30 times per problem instance and for a maximum number of evaluations equal to $E_{\max}= 2 \cdot 10^7$. All runs of local search techniques inside the memetic versions were limited to exploring $n_{\nu}=2\cdot10^6$ neighbours. This number corresponds to the maximum number of backtrack steps (fixing one entry of the incidence matrix) performed by CLS in \cite{DBLP:conf/cp2003/PrestwichCP03}. The GAs consider the equivalent number of full evaluations in each case. The number of evaluations without
improvement to trigger intensification in a Local Search method or re-starting in a population-based method is $n_\iota = n_\nu/10$. Other parameters of the GA/MA are population size \emph{popsize} $=100$, crossover and mutation probabilities $p_X=.9$ and $p_M=1/\ell$ (where $\ell=vb$ is the size of individuals) respectively, $f_\%=10\%$ and binary tournament selection of parents to be recombined. We have also considered 2 and 4 parents for recombination (i.e. $m \in \{2,4\}$) and --particularly for the MA-- $p_{LS}$ was set to 0.005. In addition, the number of cycles $\Theta$ was set to 5 in the cooperative versions. 
These parameter values were chosen because some preliminary experiments indicated that they provided a good trade-off between the computational cost and the quality of solutions attained. Note however, that most of these values were the same as those used in \cite{RuedaCF09,DBLP:journals/ijcopi/RuedaCL11}.
The versions of HC, TS and GA working on the primal model (i.e. in the binary search space) are those described in \cite{RuedaCF09}. The MAs with Gd correspond to the versions described in \cite{DBLP:journals/ijcopi/RuedaCL11}.

The dual versions of these algorithms all use the same parameters (population size, genetic operator rates, etc) as the corresponding primal versions. Versions with symmetry breaking follow the considerations described in Sect.~\ref{sec:SymmetryBreakingPrimalModel} (for the primal model-based algorithms) and Sect.~\ref{sec:BreakingSDualModel} (for the dual model-based algorithms). The combination of the two problem representation models and the possibility of breaking the symmetries gave rise to four different  scenarios, namely,  primal representation with and without symmetry breaking, and their  equivalents in the dual model.

\subsection{Basic and integrative approaches}
\label{subsect:non-coop algo experiments}
Thirty two basic and integrative algorithms have been considered, i.e. 8 local search algorithms (resulting from the two LS methods considered in this paper --HC and TS-- and the four aforementioned scenarios), 8 GAs (resulting from the four previous scenarios plus two different arities for greedy recombination), and 16 MAs (resulting from the integration of either HC or TS, for performing local improvement, in each of the previous GAs). The performance results obtained for all these metaheuristics are reported in Table \ref{tab:InsResueltas86}, which shows the number of problem instances (out of 86) that were solved in at least one run by each of the algorithms, along with the corresponding success percentage.

\begin{table}[!t]
\caption{Number (and percentage) of the instances solved by the basic and integrative  metaheuristics working alone on the set of 86 instances taken from \cite{DBLP:journals/nn/BofillGT03,DBLP:conf/cp2003/PrestwichCP03}.}
\label{tab:InsResueltas86}
\begin{center}
\scalebox{0.8}{
\begin{tabular}{|c|c|c|c|}
\hline
algorithm 			& \# (\%) 			& algorithm 		& \# (\%)\\
\hline  
Hc.B 			&  35 (40.70 \%) 	& Ts.B 			&  57 (66.28 \%) \\ 
Hc.D 			&   6 ( 6.98 \%) 		& Ts.D 			&  43 (50.00 \%) \\ 
Hc.B* 			&  25 (29.07 \%) 	& Ts.B* 			&  51 (59.30 \%) \\ 
Hc.D* 			&   3 ( 3.49 \%) 		& Ts.D* 			&  46 (53.49 \%) \\ 
GA.B.A2.Gd 		&  25 (29.07 \%) 	& GA.B.A4.Gd 		&  35 (40.70 \%) \\ 
GA.D.A2.Gd 		&  35 (40.70 \%) 	& GA.D.A4.Gd 		&  36 (41.86 \%) \\ 
GA.B*.A2.Gd 		&  38 (44.19 \%) 	& GA.B*.A4.Gd 	&  43 (50.00 \%) \\ 
GA.D*.A2.Gd 		&  28 (32.56 \%) 	& GA.D*.A4.Gd 	&  31 (36.05 \%) \\ 
MA.Hc.B.A2.Gd 	&  43 (50.00 \%) 	& MA.Ts.B.A2.Gd 	&  53 (61.63 \%) \\ 
MA.Hc.B.A4.Gd 	&  46 (53.49 \%) 	& MA.Ts.B.A4.Gd 	&  56 (65.12 \%) \\ 
MA.Hc.D.B.A2.Gd 	&  14 (16.28 \%) 	& MA.Ts.D.B.A2.Gd 	&  52 (60.47 \%) \\ 
MA.Hc.D.B.A4.Gd 	&  15 (17.44 \%) 	& MA.Ts.D.B.A4.Gd	&  52 (60.47 \%) \\ 
MA.Hc.B*.A2.Gd 	&  43 (50.00 \%) 	& MA.Ts.B*.A2.Gd 	&  53 (61.63 \%) \\ 
MA.Hc.B*.A4.Gd 	&  46 (53.49 \%) 	& 
{\bf MA.Ts.B*.A4.Gd} 	&  {\bf 59 (68.60 \%)} \\ 
MA.Hc.D*.A2.Gd 	&  10 (11.63 \%) 	& MA.Ts.D*.A2.Gd 	&  51 (59.30 \%) \\ 
MA.Hc.D*.A4.Gd 	&   9 (10.47 \%) 	& MA.Ts.D*.A4.Gd 	&  47 (54.65 \%) \\ 
\hline 
\end{tabular}
}
\end{center}
\end{table}

In general, TS variants outperform both their HC counterparts and GA versions. More specifically, the TS algorithm working on the swap neighbourhood and the primal model (i.e. Ts.B) has proven to be very efficient (this confirms the results shown in \cite{RuedaCF09}). We have also found that using symmetry breaking in the best memetic proposal (i.e. MA.Ts.B.A4.Gd) described in \cite{DBLP:journals/ijcopi/RuedaCL11} produces an improvement: MA.Ts.B*.A4.Gd solves 59 instances while only 56 are solved by MA.Ts.B.A4.Gd.

\subsubsection{A rank-based comparison}
\begin{figure}[!t]
\begin{center}
\includegraphics[width=0.9\columnwidth]{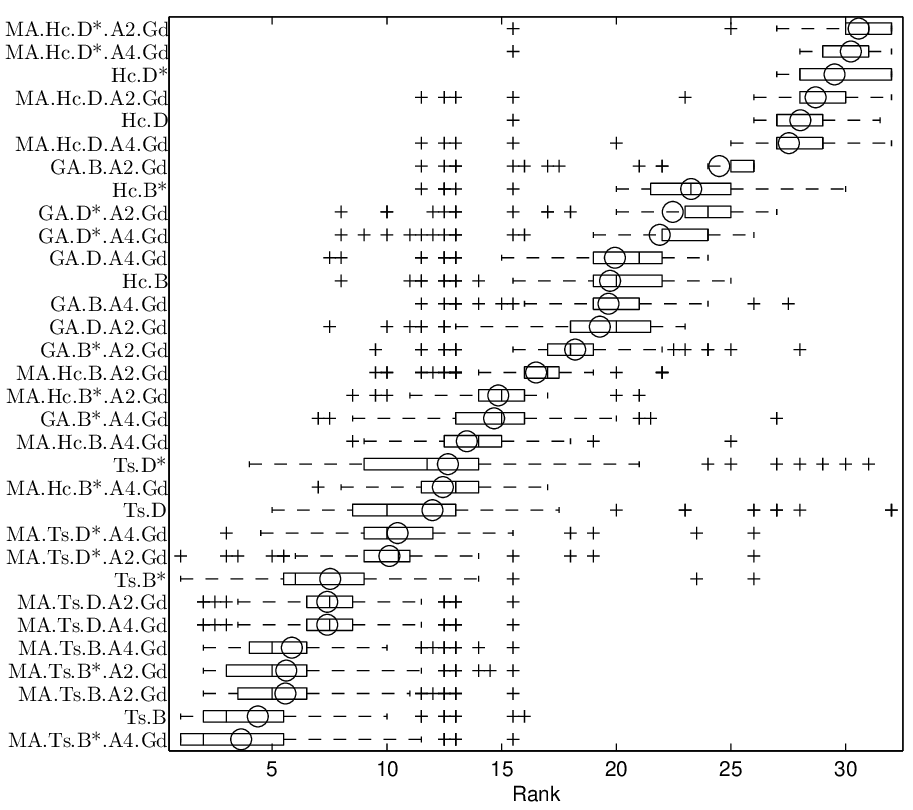}
\end{center}
\caption{Rank distribution of basic and integrative metaheuristics on the 86 instances taken from \cite{DBLP:journals/nn/BofillGT03,DBLP:conf/cp2003/PrestwichCP03}. 
}
\label{fig:TorneoTodos}
\end{figure}

Due to the high number of algorithm variants, it is not easy to compare their performances by simply inspecting the numerical tables. Therefore, we
have opted for a rank-based approach. More precisely, we have computed the rank $r_j^i$ of each algorithm $j$ on each instance $i$.  
For ranking purposes, we have used the number of solutions found for each instance (from the set of 30 runs) and employed the mean fitness to break ties.
The best algorithm is ranked first and the worst  is ranked 32nd. The distributions of these ranks are shown in Figure~\ref{fig:TorneoTodos}. At first glance, the integrative cooperative algorithms (i.e. the MA versions) perform better than non-cooperative counterparts. The results confirm that the MA with symmetry breaking (i.e. MA.Ts.B*.A4.Gd) outperforms the one without it and can now be considered the best metaheuristic for the BIBD problem. A more detailed statistical analysis indicates that there are significant differences ($\alpha=0.05$) among the different algorithms according to the Friedman test \cite{Friedman1937} and the Iman-Davenport test \cite{Iman1980}. For this reason, we carried out a post-hoc analysis using the Holm-Bonferroni test \cite{Holm1979}. As shown in Table \ref{tab:holmAlone}, MA.Ts.B*.A4.Gd is significantly better than the other algorithms, except Ts.B and the other three MAs using TS on the primal model (regardless of the use of symmetry breaking or recombination arity).

\begin{table}[!ht]
\caption{Results of the Holm-Bonferroni test on integrative approaches using \textsf{MA.Ts.B*.A4.Gd} as the control algorithm. Only the algorithms that show no significant statistical difference --at the standard level $\alpha = 0.05$-- with respect to 
the control algorithm are shown (i.e. those for which p-value $\ge \alpha/i$).}
\label{tab:holmAlone}
\begin{center}
\scalebox{0.8}{
\begin{tabular}{lcccc}
\hline
$i$ & algorithm & z-statistic  &  p-value &       $ \alpha / i $ \\ 
\hline
1 & \textsf{Ts.B} & 5.039e-001 & 3.071e-001 & 5.000e-002\\ 
2 & \textsf{MA.Ts.B.A2.Gd} & 1.345e+000 & 8.928e-002 & 2.500e-002\\ 
3 & \textsf{MA.Ts.B*.A2.Gd} & 1.374e+000 & 8.477e-002 & 1.667e-002\\ 
4 & \textsf{MA.Ts.B.A4.Gd} & 1.544e+000 & 6.125e-002 & 1.250e-002\\ 
\hline 
\end{tabular}}
\end{center}
\end{table}

\begin{figure}[!t]
\begin{center}
\includegraphics[width=0.9\columnwidth]{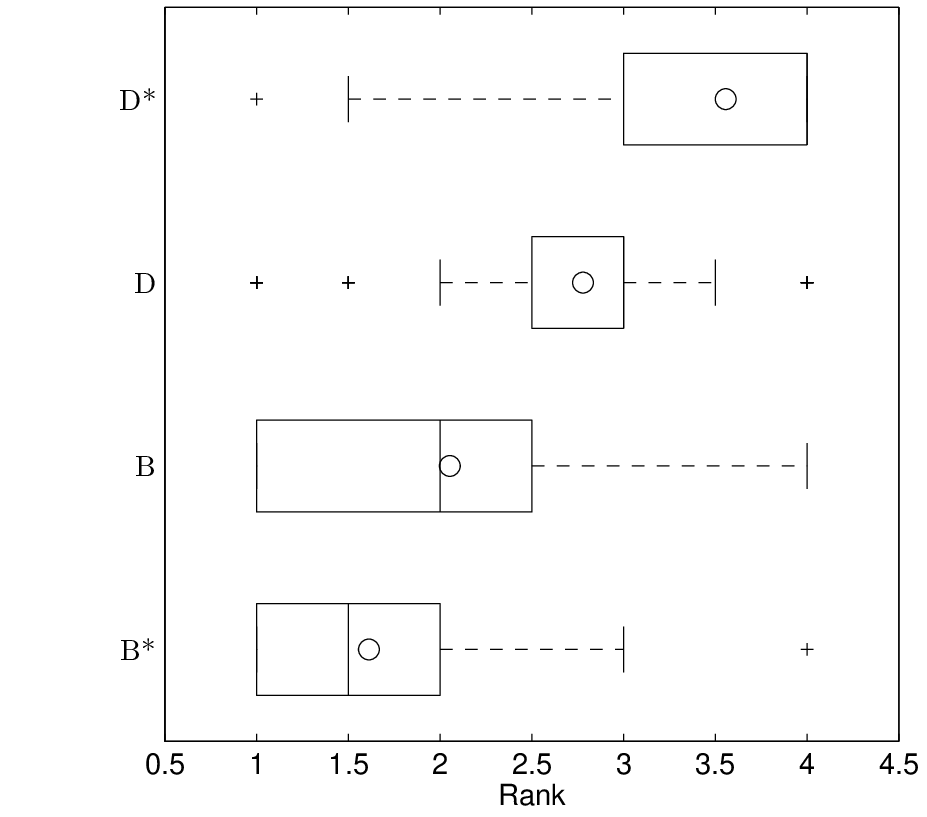}
\end{center}
\caption{Rank distribution of the four groups: \{primal, dual\}$\times$\{with symmetry-breaking, without symmetry-breaking\}.
B identifies the Binary (i.e. Primal) model without symmetry breaking, B* the Binary Model with symmetry breaking,
D the Decimal (i.e. Dual) model without symmetry breaking, and  D* the Decimal model with symmetry breaking.}
\label{fig:TorneoCSRupturaModeloALL}
\end{figure}

\subsubsection{A factor-based comparison}

Some interesting observations emerge when the data is factorised along particular dimensions. To begin with, let us consider the representation used. If we compare the algorithms operating on the primal representation with those operating on the dual representation, we observe 
a highly significant difference in favour of the primal representation (according to a Wilcoxon signed rank test \cite{Lehman1998}, p-value $\approx 0$). This result also holds for each group of techniques (LS, GA, MA) when analysed separately (p-values always below $0.0001$).  Now, if an analogous analysis with regard to the use of symmetry breaking (SB) is performed considering all algorithms, it turns out that not using SB is significantly better (again, p-value $\approx 0$) as well as for LS and MAs.

A joint analysis of representation and use of symmetry breaking was also executed splitting the data into four groups: \{primal, dual\}$\times$\{with symmetry breaking, without symmetry breaking\} (i.e. B, D, D and D*). As can be seen in Figure \ref{fig:TorneoCSRupturaModeloALL}, using the primal model with symmetry breaking yields the best global rank, and significantly outperforms the other combinations. The Friedman test and the Iman-Davenport test confirm that there are significant differences (at the standard $\alpha=0.05$) among the different groups. Results of the Holm-Bonferroni test, applied as a post-hoc analysis and using \textsf{B$^*$} as control algorithm, indicates that the control algorithm shows significant statistical differences ($\alpha = 0.05$) with the other algorithms (i.e. in all the cases, p-value $< \alpha/i$).

Note, we have stated above that --globally speaking-- the use of symmetry breaking (SB) in the basic metaheuristics working as standalone methods is not recommendable. However, we emphasise that we have utilised just one type of symmetry breaking, called variable reduction (VR) in \cite{G2016117}. Other types of symmetry breaking --such as, e.g. lexicographical ordering-- may have more added value. In any case,
we have detected that, when our SB proposal is used by some population-based methods, the performance improves. So, the best memetic algorithm (i.e. MA.Ts.B*.A4.Gd) as well as the two best GA versions (i.e. GA.B*.A2.Gd and GA.B*.A4.Gd) use symmetry breaking. Moreover, SB seems to work well when combined with the primal model but not with the dual model. This demonstrates that each representation and symmetry-breaking approach provides a different angle to the search process. This raises interesting prospects on their joint use in cooperative models, whose performance is analysed next.

\subsection{Cooperative approaches}
\label{subsect:coop algo experiments}
This section evaluates the performance of a number of different cooperative algorithms (instantiated from the scheme shown in Algorithm \ref{alg:model} described in Sect. \ref{Sec:EsquemasCooperativos}). The idea is to harness the synergy between the metaheuristics when these work in cooperation. We have considered the three topologies proposed with a number $n$ of agents between 2 and 5 (following \cite{Amaya2011MC}), and a number of cycles $\Theta=5$ (we set this value based on preliminary experiments  with values of $\Theta \in \{5,10,15 \}$).

The cooperative models considered are variations of the template (see Sect. \ref{sec:notation}) $\mathbf{T}n(p a,q b)$MR, where $p+q=n$ and $a,b\in{\cal A}$  for a certain collection ${\cal A}$ that contains two types of agents. We have considered  the following four collections:

\begin{itemize}
\item ${\cal A}_1 =$ \{Ts.B,MA.Ts.B*.A4.Gd\}
\item ${\cal A}_2 =$\{Ts.B,MA.Ts.B.A2.Gd\}
\item ${\cal A}_3 =$\{Ts.B,MA.Ts.D.A4.Gd\} 
\item ${\cal A}_4 =$\{Ts.B,MA.Ts.D*.A2.Gd\}
\end{itemize}

The algorithms in these collections have been picked due to their good individual performances according to Table \ref{tab:holmAlone}. Ts.B is the best basic technique whereas  the other four algorithms  in these collections represent the best integrative methods in the domains B*, B, D and D*. Moreover, each collection represents a form of combining algorithms:  ${\cal A}_1$ represents the cooperation of a model (in this case, the Binary representation) with and without symmetry breaking (i.e. B-B*),  ${\cal A}_2$ represents the cooperation of techniques working in the same computation domains with no symmetry breaking (i.e. B-B),  ${\cal A}_3$ represents the cooperation of methods working in distinct computation domains with no symmetry breaking (i.e. B-D), and ${\cal A}_4$ the scheme in which methods working on distinct computation domains  with distinct policies for symmetry breaking are cooperating (i.e. B-D*). Considering all possible combinations of topology, number of agents and migration/reception policies, a total of 288 algorithmic variants were created. The value 288 results from combining: (1) two different representations (i.e. primal, dual), (2) two distinct forms of managing the problem solving (i.e. with or without symmetry breaking), (3) three distinct communication topologies (i.e. \textsc{Ring}, \textsc{Rand} or \textsc{Broadcast}), (4) six options for  migration/reception policies, and (5) 4 collections of algorithms to load the agents in the 4-agent scheme  (i.e. collections ${\cal A}_1$,${\cal A}_2$,${\cal A}_3$ and ${\cal A}_4$).

Due to the large number of variants, we consider a reduced set of 29 instances to evaluate the performance of the algorithms. More specifically, we use the 29 instances that could not be solved by  the (non-constructive) metaheuristics methods mentioned in Sect.~\ref{sect:related work}\footnote{Note that an MA with Gd proposed in \cite{DBLP:journals/ijcopi/RuedaCL11} was able to  solve 63 of the 86 instances but it required a substantially larger number of evaluations compared to the same algorithm that solved 57 instances as reported in \cite{RuedaCF09}.}.
These 29 hard problem instances are shown in Table~\ref{tab:29Instancias}. 
 
\begin{table}[!t]
\caption{The 29 problem instances considered hard from 
the 86 instances in \cite{DBLP:journals/nn/BofillGT03,DBLP:conf/cp2003/PrestwichCP03}. The first column is the instance label assigned in \cite{DBLP:conf/cp2003/PrestwichCP03}, columns 2--6 present the instance parameters, and column 7 gives an indication of the size of the instance.}
\label{tab:29Instancias}
\begin{multicols}{2}

\begin{center}
\scalebox{0.7}{
\begin{tabular}{|c|ccccc|c|}
\hline
ID& $v$ &  $b$ &  $r$ & $k$  & $\lambda$  & $vb$  \\
\hline
21&14&26&13&7&6&    364\\	
27&15&30&14&7&6&    450 \\	
28&16&30&15&8&7&    480 \\	
33&16&32&12&6&4&    512 \\	
34&15&35&14&6&5&    525 \\	
39&17&34&16&8&7&    578 \\  
43&18&34&17&9&8&    612 \\
44&25&25&9&9&3&    625 \\
46&21&30&10&7&3&    630 \\
48&16&40&15&6&5&    640 \\
50&15&45&21&7&9&     675 \\
54&19&38&18&9&8&      722 \\
56&22&33&12&8&4&     726 \\
57&14&52&26&7&12&    780 \\
58&27&27&13&13&6&   729 \\
\hline
\end{tabular}}
\end{center}

\begin{center}
\scalebox{0.7}{
\begin{tabular}{|c|ccccc|c|}
\hline
ID& $v$ &  $b$ &  $r$ & $k$  & $\lambda$  & $vb$  \\
\hline
59&21&35&15&9&6&     735 \\
62&20&38&19&10&9&   760 \\
63&16&48&18&6&6&    768 \\
70&21&42&10&5&2&     882 \\
71&21&42&12&6&3&     882 \\
72&21&42&20&10&9&    882 \\
73&16&56&21&6&7&     896 \\
76&18&51&17&6&5&      918 \\
77&22&42&21&11&10&   924 \\
80&16&60&30&8&14&      960 \\
82&31&31&10&10&3&     961 \\
83&31&31&15&15&7&     961 \\
85&22&44&14&7&4&       968 \\
86&25&40&16&10&6&     1000 \\
\hline
\end{tabular}}
\end{center}
\end{multicols}
\end{table}

\subsubsection{Analysis of Design Factors}
\label{sec:three-dimensional}
Given the large number of algorithms, it is useful to factorise the analysis along different dimensions corresponding to different design decisions regarding the number of agents involved, their topology, or the communication policy. Let us start by considering the six combinations (i.e. DD, DR, DW, RD, RR and RW) used. Comparing the results of these six policies on any single algorithmic variant. More precisely, for each of the 288/6=48 variants (resulting from different combinations of topology, number of agents and individual algorithms used), we ranked the six migration/reception (M/R) policies according to the number of optimal solutions found (using the mean fitness to break ties). The best M/R policy is RD (random selection of migrants, replacement for diversity). This is further confirmed by a Friedman and Iman-Davenport tests (at the standard level of $\alpha=0.05$) which indicated that there are statistically significant differences among policies, and by a Holm-Bonferroni test that showed that RD is significantly better than the other policies.

Next, we consider an analysis along the topology axis. In this case, the three topologies are ranked on 288/3=96 algorithmic variants each. As indicated by the Friedman and Iman-Davenport tests ($\alpha=0.05$) for all topologies, there is a statistically significant difference in this case. Indeed, the \textsc{Broadcast} topology stands out from the others, as confirmed by the Holm-Bonferroni employing \textsc{Broadcast} as control algorithm. 
The larger exchange of information among agents in this topology might be the cause.

 We now turn our attention to the number of agents used in the cooperative model. We have considered $n\in[2,5]$ and these four values are ranked across 288/4=72 algorithmic variants each.  Both the Friedman test and the Iman-Davenport test indicate that there are significant differences among the different values, so we conducted  the Holm-Bonferroni test using $n=2$ (the best ranked value) as the control algorithm.  The result was that the difference is significant against the remaining values of $n$. The number of agents is therefore a factor that exerts a significant influence on the performance of the cooperative model. 
In our experiments, the algorithms that employ 2 agents perform better than those that employ more agents. In addition,  the algorithms that employ 2 or 3 agents perform, in general, better than those based on 4-5 agents. 
 
\subsubsection{Analysis of Top Performing Models}
\label{sec:Top41}
We now focus on the most effective cooperative models. More specifically, we consider all cooperative models that were able to solve at least 9 (hard) problem instances (out of the 29 mentioned above) in at least one run. This set of algorithms is composed of the 41  variants shown in Table \ref{tab:InsResuTop41} together with the number of problem instances solved and the corresponding success percentage. 

Note first that  there is a predominance of the collection ${\cal A}_2$. The collaboration of techniques working in the same computation domain (in this case, Binary encoding) is present 
in 38 of the 41 algorithms. Good results are also provided by 3 other cooperative algorithms in which the collaboration of methods is based on the collection ${\cal A}_3$, that is to say, 
techniques working collaboratively on the primal and dual models. However, encouraging cooperation between methods that manage symmetry breaking and other techniques that do not consider symmetry issues does not seem to be helpful.
Note however that this statement should not be generalised to any symmetry breaking. 
Specifically, for the case considered here, 
cooperation between agents working in B-B* or B-D* (i.e. those based on collections ${\cal A}_1$ or ${\cal A}_4$) is not advisable. Both the Friedman test and the Iman-Davenport test (with $\alpha=0.05$) indicate that there are significant differences among the different collections of algorithms. Moreover, the Holm-Bonferroni test using ${\cal A}_2$ as control algorithm confirms that it is statistically significant from the other collections.

\begin{table}[!t]
\caption{(Central column) Number (\#) and percentage (\%)  of instances from Table \ref{tab:29Instancias}  solved by those cooperative algorithms (identified in the first column) that were successful in at least 9 (hard) problem instances. Right column displays the collection of algorithms that collaborate in the cooperative search.}
\label{tab:InsResuTop41}
\begin{center}
\scalebox{0.7}{
\begin{tabular}{|c|c|c|}
\hline
Algorithms & \# (\%) & Collection\\
\hline
Bc2(Ts.B,MA.Ts.B.A2.Gd)DD &   9 (31.03 \%)			&  ${\cal A}_2$\\ 
Bc2(Ts.B,MA.Ts.B.A2.Gd)DR &   9 (31.03 \%)			&  ${\cal A}_2$\\ 
Bc2(Ts.B,MA.Ts.B.A2.Gd)DW &  12 (41.38 \%)			&  ${\cal A}_2$\\
Bc2(Ts.B,MA.Ts.B.A2.Gd)RD &  11 (37.93 \%)			&  ${\cal A}_2$\\
Bc2(Ts.B,MA.Ts.B.A2.Gd)RR &  12 (41.38 \%)			&  ${\cal A}_2$\\
Bc2(Ts.B,MA.Ts.B.A2.Gd)RW &   9 (31.03 \%)			&  ${\cal A}_2$\\
Ra2(Ts.B,MA.Ts.B.A2.Gd)DD &  12 (41.38 \%)			&  ${\cal A}_2$\\
Ra2(Ts.B,MA.Ts.B.A2.Gd)DR &  10 (34.48 \%)			&  ${\cal A}_2$\\
Ra2(Ts.B,MA.Ts.B.A2.Gd)DW &  11 (37.93 \%)			&  ${\cal A}_2$\\
Ra2(Ts.B,MA.Ts.B.A2.Gd)RD &  10 (34.48 \%)			&  ${\cal A}_2$\\
Ra2(Ts.B,MA.Ts.B.A2.Gd)RR &  12 (41.38 \%)			&  ${\cal A}_2$\\
Ra2(Ts.B,MA.Ts.B.A2.Gd)RW &  12 (41.38 \%)			&  ${\cal A}_2$\\
Ri2(Ts.B,MA.Ts.B.A2.Gd)DD &   9 (31.03 \%)			&  ${\cal A}_2$\\
{{\bf{Ri2(Ts.B,MA.Ts.B.A2.Gd)DR}}} &  {{\bf{13 (44.83 \%)}}}			&  ${\cal A}_2$\\
Ri2(Ts.B,MA.Ts.B.A2.Gd)RD &  10 (34.48 \%)			&  ${\cal A}_2$\\
Ri2(Ts.B,MA.Ts.B.A2.Gd)RR &  10 (34.48 \%)			&  ${\cal A}_2$\\
Ri2(Ts.B,MA.Ts.B.A2.Gd)RW &  11 (37.93 \%)			&  ${\cal A}_2$\\
Bc3(2Ts.B,MA.Ts.B.A2.Gd)DD &   9 (31.03 \%)			&  ${\cal A}_2$\\
Bc3(2Ts.B,MA.Ts.B.A2.Gd)DR &   9 (31.03 \%)			&  ${\cal A}_2$\\ 
Bc3(2Ts.B,MA.Ts.B.A2.Gd)DW &   9 (31.03 \%)			&  ${\cal A}_2$\\
Bc3(2Ts.B,MA.Ts.B.A2.Gd)RD &  11 (37.93 \%)			&  ${\cal A}_2$\\
Bc3(2Ts.B,MA.Ts.B.A2.Gd)RR &   9 (31.03 \%)			&  ${\cal A}_2$\\
Bc3(2Ts.B,MA.Ts.B.A2.Gd)RW &   9 (31.03 \%)			&  ${\cal A}_2$\\
Ra3(2Ts.B,MA.Ts.B.A2.Gd)DD &   9 (31.03 \%)			&  ${\cal A}_2$\\
{{\bf{Ra3(2Ts.B,MA.Ts.B.A2.Gd)RD}}} &  {{\bf{13 (44.83 \%)}}}			&  ${\cal A}_2$\\
Ra3(2Ts.B,MA.Ts.B.A2.Gd)RR &   9 (31.03 \%)			&  ${\cal A}_2$\\
Ri3(2Ts.B,MA.Ts.B.A2.Gd)DD &   9 (31.03 \%)			&  ${\cal A}_2$\\
Ri3(2Ts.B,MA.Ts.B.A2.Gd)DR &  10 (34.48 \%)			&  ${\cal A}_2$\\
Ri3(2Ts.B,MA.Ts.B.A2.Gd)RD &  10 (34.48 \%)			&  ${\cal A}_2$\\
Ri3(2Ts.B,MA.Ts.B.A2.Gd)RR &   9 (31.03 \%)			&  ${\cal A}_2$\\
Bc4(2Ts.B,2MA.Ts.B.A2.Gd)DR &  10 (34.48 \%)			&  ${\cal A}_2$\\
Bc4(2Ts.B,2MA.Ts.B.A2.Gd)RR &   9 (31.03 \%)			&  ${\cal A}_2$\\
Ra4(2Ts.B,2MA.Ts.B.A2.Gd)DW &   9 (31.03 \%)			&  ${\cal A}_2$\\
Bc5(3Ts.B,2MA.Ts.B.A2.Gd)DD &   9 (31.03 \%)			&  ${\cal A}_2$\\
Bc5(3Ts.B,2MA.Ts.B.A2.Gd)DR &  10 (34.48 \%)			&  ${\cal A}_2$\\
Bc5(3Ts.B,2MA.Ts.B.A2.Gd)RD &   9 (31.03 \%)			&  ${\cal A}_2$\\
Bc5(3Ts.B,2MA.Ts.B.A2.Gd)RW &   9 (31.03 \%)			&  ${\cal A}_2$\\
Ra5(3Ts.B,2MA.Ts.B.A2.Gd)DW &   9 (31.03 \%)			&  ${\cal A}_2$\\
Bc2(Ts.B,MA.Ts.D.A4.Gd)RD &   9 (31.03 \%)			&  ${\cal A}_3$\\
Ra5(3Ts.B,2MA.Ts.D.A4.Gd)RD &   9 (31.03 \%)			&  ${\cal A}_3$\\
Ri5(3Ts.B,2MA.Ts.D.A4.Gd)RD &   9 (31.03 \%)			&  ${\cal A}_3$\\
\hline 
\end{tabular}}
\end{center}
\end{table}

If we check the relative frequency of each particular design parameter among these 41 algorithmic models, we obtain the results shown in Table \ref{tab:NumPorcxCriterioTop41}.  The results are consistent with the statistical analysis from the previous section. Thus, we can see that the RD policy for migration/reception performs best. Regarding topology, \textsc{Broadcast} is present in more than half of the models. Finally, a low number of agents 
(i.e. 2 or 3 agents)
seems to offer a good performance more frequently. 

\begin{table}[!t]
\caption{Relative frequency of each particular design parameter among the selected cooperative algorithms in Table \ref{tab:InsResuTop41}. Left column indicates the combination MR of migration(M)/reception(R) policies where  M, R $\in \{\textsc{Random}$ (R), $\textsc{Diverse}$ (D), $\textsc{Worst}$ (W) as explained in Sect. \ref{sec:notation}. Central column shows the communication topology where Bc = \textsc{Broadcast}, Ra = \textsc{Random}, and Ri = \textsc{Ring}. Right column refers to the number of agents in the algorithm. }
\label{tab:NumPorcxCriterioTop41}
\begin{center}
\scalebox{0.7}{
\begin{tabular}{crrlcrrlcrr}
\hline
\multicolumn{3}{c}{M/R Policy} & ~\hspace{2mm}~ & \multicolumn{3}{c}{Topology} & ~\hspace{2mm}~ & \multicolumn{3}{c}{Number of agents} \\
\cline{1-3}\cline{5-7}\cline{9-11}
DD 	& ~7 &(17.07 \%) 	&& Bc	& 19 &(46.34 \%) && $n=2$	& 18 &(43.90 \%) \\
DR 	& ~7 &(17.07 \%) 	&& Ra	& 12 &(29.27 \%) && $n=3$	& 13 &(31.71 \%) \\
DW 	& ~5 &(12.19 \%)	&& Ri	& 10 &(24.39 \%) && $n=4$	& ~3 &(~7.32 \%) \\
RD	& 10 &(24.39 \%)	&&		&	&		     && $n=5$	& ~7 &(17.07 \%)\\
RR 	& ~7 &(17.07 \%)	&& \\
RW	& ~5 &(12.19 \%)	&&\\
\hline
\end{tabular}}
\end{center}
\end{table}

\begin{figure}[!t]
\begin{center}
\includegraphics[width=0.9\columnwidth]{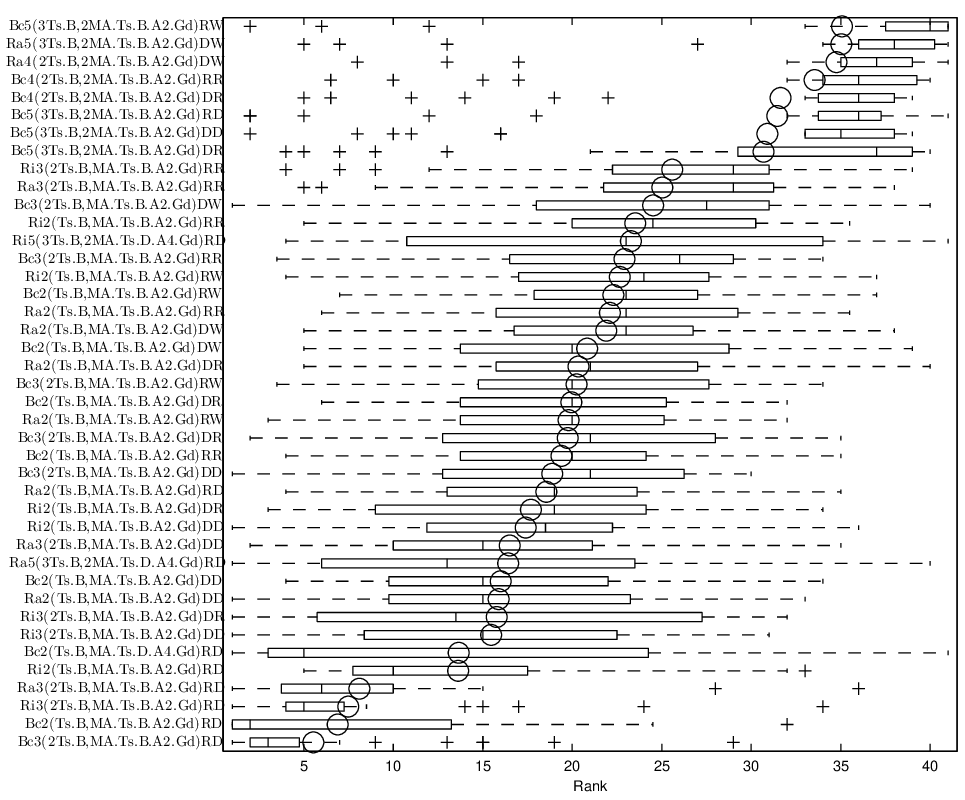}
\end{center}
\caption{Rank distribution of selected cooperative algorithms from Table \ref{tab:InsResuTop41}.}
\label{fig:TorneoCooperandoTop41}
\end{figure}

Figure \ref{fig:TorneoCooperandoTop41} shows the rank distribution for the 41 models. The Friedman and Iman-Davenport tests indicate significant differences in their ranks ($\alpha=0.05$). Subsequently, the Holm-Bonferroni test, using as the control algorithm Bc3(2Ts.B,MA.Ts.B.A2.Gd)RD, shows that there is no statistical difference between the first four cooperative algorithms, but there is a statistically significant difference with the others (see results in Table \ref{tab:holmCooperaTop41}).

\begin{table}[!ht]
\caption{Results of the Holm-Bonferroni test, for cooperative algorithms using \textsf{Bc3(2Ts.B,MA.Ts.B.A2.Gd)RD} -- i.e. the algorithm with the best mean rank according to Figure \ref{fig:TorneoCooperandoTop41}-- as the control algorithm. Only the algorithms that show no significant difference --$\alpha = 0.05$-- to the control algorithm are shown (i.e. those for which p-value $\ge \alpha/i$).}
\label{tab:holmCooperaTop41}
\begin{center}
\scalebox{0.7}{
\begin{tabular}{lcccc}
\hline
$i$ & algorithm & z-statistic  &  p-value &       $ \alpha / i $ \\ 
\hline
1 & \textsf{Bc2(Ts.B,MA.Ts.B.A2.Gd)RD} & 4.330e-001 & 3.325e-001 & 5.000e-002\\ 
2 & \textsf{Ri3(2Ts.B,MA.Ts.B.A2.Gd)RD} & 6.193e-001 & 2.679e-001 & 2.500e-002\\ 
3 & \textsf{Ra3(2Ts.B,MA.Ts.B.A2.Gd)RD} & 8.166e-001 & 2.071e-001 & 1.667e-002\\ 
\hline 
\end{tabular}}
\end{center}
\end{table}

\subsection{Discussion: Cooperation vs. Integration}
\label{sec:analysis}
The high-level distinction between integration and cooperation is well known in the literature and has been used to taxonomise the hybridisation of search algorithms (see e.g. \cite{Puchinger2005}).
This section discusses and compares the performance of our cooperative and integrative approaches.
Note that we have first compared a collection of 32 integrative metaheuristics on the 86 classical instances of the problem. An interesting result is that the best metaheuristic approach described previously in the literature, namely a memetic algorithm, was improved by adding symmetry breaking. This new algorithm solved 59 out of 86 instances (i.e. 68.60$\%$ effectiveness). Despite this improvement, there were 27 instances that could not be solved. These instances are part of a subset of 29 instances considered difficult to solve. To tackle these harder instances, we proposed a number of cooperative algorithms. Two of these could solve 13 of the hard instances, and, overall, 14 different instances were solved by some of the cooperative methods, namely, instances 21, 27, 28, 33, 34, 39, 44, 48, 50, 57, 58, 64, 73, and 76 in Table~\ref{tab:29Instancias} (the precise set of problems solved by each method is shown in \cite{datainbriefSubmitted}). In summary, this means that we could solve 71 out of 86 instances (i.e. $82.56\%$ effectiveness) by either method (i.e. 57 instances from the original set of 86 plus 14 hard ones). It is important to underline that a large number of cooperative algorithms were successful for many problem instances where simpler metaheuristics did not succeed. 
Note that this result, which shows the advantages of the cooperative techniques with respect to their underlying components  (i.e. the algorithms that compose the collaboration) working alone, is consistent  with other studies conducted for problems other than the BIBD problem. For instance, the results are similar to those obtained in \cite{Amaya2011} in the context of the tool switching problem, and our findings are consistent with the conclusions of this paper.
Needless to say, not all integrative methods (or cooperative methods) perform in the same way. Indeed, the crux of the whole matter with regards to effectiveness is often (1) the balance between exploration and exploitation and (2) the search overhead in which composite methods incur. These two issues are influenced by different factors and in this sense, one of the contributions of the work presented here,
with respect to the conclusions regarding integration and cooperation of algorithms,  is to calibrate them in this particular domain. Another point of novelty is to shine a spotlight on two more dimensions (in addition to the methods used in integrative hybrid and the cooperative schemes defined with them), namely symmetry breaking and problem primal/dual formulations, which to the best of our knowledge has not been explored before in this context. 

Additionally, we highlight the importance of the policies for migrating and accepting solutions from the agents in the metaheuristic network for the performance of the cooperative algorithms. 
In our experiments, using the \textsc{Worst} policy for migration of candidates deteriorated the performance of the algorithm. In general, the combination RD has a positive effect on cooperation as the first six algorithms from the top-ten cooperative algorithms are based on it. Also note that we have executed a large number of experiments, and considered other combinations and algorithms, many of which are not reported here to avoid clutter, given that their performance was poor. 

Another very interesting result that can be extracted from our experiments is that cooperation performs better when all the connected algorithms are working in the same computation domain and using the same problem formulation. 

\begin{table}[!t]
\caption{Running time comparison between MA=$Ma.Ts.B*.A4.Gd$
and Coop=$Ra3(2Ts.B,Ma.Ts.B.A2.Gd)RD$.
The two problem instances $I_1 = \langle14,26,13,7,6 \rangle$ and $I_2 = \langle 25,25,9,9,3 \rangle$ 
that were solved by both methods are selected as benchmarks for comparison.  $T_{max}/T_{min}$ show the maximum/minimum time (in seconds) consumed by the best execution of each technique to find a solution. Last column displays the improvement of the cooperative method with respect to the MA.}
	\label{tab:Running time comparison 2 instances}
	\begin{center}
		\scalebox{0.8}{
		\begin{tabular}{ccccccc} \cline{2-5}
			          &\multicolumn{4}{ c }{Instances}  & &  \\ \cline{2-7}
			          &\multicolumn{2}{ c}{$I_1$} &  \multicolumn{2}{ c}{$I_2$} &  \multicolumn{2}{ c }{Coop/MA(\%)}   \\ \hline
			\multicolumn{1}{ c }{Time(s)}& MA  &  Coop &  MA  &  Coop & \multicolumn{1}{ c}{$I_1$} &  \multicolumn{1}{ c}{$I_2$} \\ \hline
			\multicolumn{1}{ c }{$T_{min}$} & 23.00 &  0.02 &  20.08  & 0.03 & \multicolumn{1}{ c}{1150} &  \multicolumn{1}{ c}{669} \\ \hline
			\multicolumn{1}{ c }{$T_{max}$} & 5754.78 &  24.67 &  9317.03  & 34.91 & \multicolumn{1}{ c}{233}  &  \multicolumn{1}{ c}{266} \\ \hline			
\end{tabular}
}
\end{center}
\end{table}

With respect to running times, we have compared the best integrative approach (i.e. $Ma.Ts.B*.A4.Gd$, which could only solve 2 of the 29 hard problem instances shown in Table \ref{tab:29Instancias}),
 with one of our best cooperative methods (i.e. $Ra3(2Ts.B,Ma.Ts.B.A2.Gd)RD$, which solved 13 of the 29 hard instances).
To be fair, we only considered the two instances that were solved by both methods. The results are shown in Table \ref{tab:Running time comparison 2 instances}. The memetic algorithm is noticeably much slower than the cooperative one. This must be due to some kind of synergy between the agents.

\section{Conclusions and future work}
\label{sec:conclusions}

This paper has dealt with the generation BIBDs, a difficult combinatorial problem, using metaheuristic techniques. In previous work, we tackled this problem by means of local search algorithms, genetic algorithms and hybrid techniques, always using  a binary genotypical space and a classical (primal) formulation of the problem. In this paper, we have proposed a number of alternative approaches to deal with this problem. First, we have defined a (novel) dual problem formulation with a natural representation in the integer domain. All heuristics proposed in previous work have been adapted to this novel formulation. In addition, based on the highly symmetrical nature of the problem, we have considered basic symmetry-breaking procedures to reduce the search space of the problem in both encodings, i.e. primal (binary) and dual (integer). The advantage of using these  procedures is highly dependent on other design decisions but the best approach has been shown to be a memetic algorithm using symmetry breaking on the primal representation, outperforming the previous best known metaheuristics for this problem. 

Despite these good results, integrative metaheuristics were somewhat limited for harder problem instances in the test suite. We have therefore also proposed a scheme to instantiate cooperative models that combine the integrative algorithms. This scheme is based on a spatial topology and policies for exchanging solutions, and allows a large number of different instantiations. We have considered three different topologies (to define the flow of information among algorithms), a varying number of connecting agents in these topologies, three different migration procedures (for selecting solutions to be sent), and three different acceptance criteria (for handling incoming solutions). All these  factors affect the performance of the algorithms. 
For a better understanding of our cooperative algorithms, we have also conducted an analysis of the influence of some of these factors. The conclusions extracted can help design better cooperative algorithms. 
In addition, some of our cooperative proposals can be considered at present, as state-of-the-art metaheuristic methods for handling BIBDs.

Moreover, the conclusions extracted from our use of primal/dual encodings, symmetry breaking  and their combination in a cooperative model may guide the design of other cooperative algorithms to handle other combinatorial problems with symmetries. In this sense, the work presented here may also be viewed as a methodology to address combinatorial problems with symmetries in a general way, that is to say, firstly designing basic metaheuristics, secondly applying hybrid methods constructed from them (i.e. memetic algorithms), then considering alternative formulation/encodings of the problem by adjusting the basic and integrative methods to these, and finally connecting all the previous techniques using cooperative schemes.

The work described here has focused on solving BIBDs. However, we believe that a large part of our work can be generalised to solve any symmetrical combinatorial problem. For this reason, our cooperative  scheme is general and so does not depend on specific algorithms, but rather on the design factors that have been analysed here. In this sense,  the metaheuristics (especially the cooperative versions) described should not be applied directly to handle open problem instances in design theory. To cope with open instances, our proposals should be tailored to the problem at hand. This basically means that we should craft our metaheuristics to suit the particular characteristics of the open instance with the aim of exploiting the available structure as much as possible. This requires not only making decisions about design factors (e.g. nature of agents, topology of the cooperative system, and migration/reception policies), but also incorporating specific knowledge about the problem instance.

In future research, we plan to deepen the study of the intensification/diversification balance of the algorithm, aiming to improve its performance for the hardest problem instances. An intermediate goal would be to endow the MAs with self-adaptive capabilities \cite{Smith2012} to enhance their search capabilities. This topic can also be explored in cooperative models with many defining parameters, which could be adjusted while running in response to the state of the search.
In addition, asymmetric representations have been shown to be effective in solving other combinatorial problems (see \cite{JANS20131132,G2016117}) and might be used in the formulation of the dual model as well.





\newcommand{\noopsort}[1]{} \newcommand{\printfirst}[2]{#1}
  \newcommand{\singleletter}[1]{#1} \newcommand{\switchargs}[2]{#2#1}








\begin{thebibliography}{55}
\expandafter\ifx\csname natexlab\endcsname\relax\def\natexlab#1{#1}\fi
\providecommand{\bibinfo}[2]{#2}
\ifx\xfnm\relax \def\xfnm[#1]{\unskip,\space#1}\fi
\bibitem[{Colbourn and Dinitz(2010)}]{colbourn2010crc}
\bibinfo{editor}{C.~Colbourn}, \bibinfo{editor}{J.~Dinitz} (Eds.),
  \bibinfo{title}{CRC Handbook of Combinatorial Designs}, Discrete Mathematics
  and Its Applications, \bibinfo{publisher}{CRC Press}, \bibinfo{year}{2010}.
\bibitem[{Hinkelmann and Kempthorne(2005)}]{PartiallyHinkelmann2005}
\bibinfo{author}{K.~Hinkelmann}, \bibinfo{author}{O.~Kempthorne},
  \bibinfo{title}{Partially Balanced Incomplete Block Designs},
  \bibinfo{publisher}{John Wiley \& Sons, Inc.}, pp. \bibinfo{pages}{119--157}.
\bibitem[{Ariel and Farrington(2014)}]{Bruinsma2014}
\bibinfo{author}{B.~Ariel}, \bibinfo{author}{D.~P. Farrington},
\newblock \bibinfo{title}{Randomized block designs},
\newblock in: \bibinfo{editor}{G.~Bruinsma}, \bibinfo{editor}{D.~Weisburd}
  (Eds.), \bibinfo{booktitle}{Encyclopedia of Criminology and Criminal
  Justice}, \bibinfo{publisher}{Springer New York}, \bibinfo{year}{2014}, pp.
  \bibinfo{pages}{4273--4283}.
\bibitem[{Buratti(1999)}]{Buratti1999103}
\bibinfo{author}{M.~Buratti},
\newblock \bibinfo{title}{Pairwise balanced designs from finite fields},
\newblock \bibinfo{journal}{Discrete Mathematics} \bibinfo{volume}{208--209}
  (\bibinfo{year}{1999}) \bibinfo{pages}{103--117}.
\bibitem[{Cheng(2014)}]{Ching-Shui2014}
\bibinfo{author}{C.-S. Cheng},
\newblock \bibinfo{title}{Regular graph designs},
\newblock in: \bibinfo{booktitle}{Encyclopedia of Statistical Sciences},
  \bibinfo{publisher}{John Wiley \& Sons, Ltd}, \bibinfo{edition}{11} edition,
  \bibinfo{year}{2014}.
\bibitem[{Fang and Lin(2008)}]{Tsubaki2008}
\bibinfo{author}{K.-T. Fang}, \bibinfo{author}{D.~Lin},
\newblock \bibinfo{title}{Uniform design in computer and physical experiments},
\newblock in: \bibinfo{editor}{H.~Tsubaki}, \bibinfo{editor}{S.~Yamada},
  \bibinfo{editor}{K.~Nishina} (Eds.), \bibinfo{booktitle}{The Grammar of
  Technology Development}, \bibinfo{publisher}{Springer Japan},
  \bibinfo{year}{2008}, pp. \bibinfo{pages}{105--125}.
\bibitem[{van Lint and Wilson(2001)}]{Lint+:CCombinatorics1992}
\bibinfo{author}{J.~van Lint}, \bibinfo{author}{R.~Wilson}, \bibinfo{title}{A
  Course in Combinatorics}, \bibinfo{publisher}{Cambridge University Press},
  \bibinfo{year}{2001}.
\bibitem[{Mead(1990)}]{Mead:DoE1993}
\bibinfo{author}{R.~Mead}, \bibinfo{title}{The Design of Experiments:
  Statistical Principles for Practical Applications},
  \bibinfo{publisher}{Cambridge University Press}, \bibinfo{year}{1990}.
\bibitem[{Buratti(1999)}]{buratti:cryptography1999}
\bibinfo{author}{M.~Buratti},
\newblock \bibinfo{title}{Some (17q, 17, 2) and (25q, 25, 3){BIBD}
  constructions},
\newblock \bibinfo{journal}{Designs, Codes and Cryptography}
  \bibinfo{volume}{16} (\bibinfo{year}{1999}) \bibinfo{pages}{117--120}.
\bibitem[{Lan et~al.(2008)Lan, Tai, Lin, Memari, and
  Honary}]{DBLP:journals/tcom/LanTLMH08}
\bibinfo{author}{L.~Lan}, \bibinfo{author}{Y.~Y. Tai},
  \bibinfo{author}{S.~Lin}, \bibinfo{author}{B.~Memari},
  \bibinfo{author}{B.~Honary},
\newblock \bibinfo{title}{New constructions of quasi-cyclic {LDPC} codes based
  on special classes of {BIDB}s for the {AWGN} and binary erasure channels},
\newblock \bibinfo{journal}{IEEE Transactions on Communications}
  \bibinfo{volume}{56} (\bibinfo{year}{2008}) \bibinfo{pages}{39--48}.
\bibitem[{dos Santos~Navarro et~al.(2014)dos Santos~Navarro, Minim, da~Silva,
  Simiqueli, Della~Lucia, Minim et~al.}]{dos2014balanced}
\bibinfo{author}{R.~d.~C. dos Santos~Navarro}, \bibinfo{author}{V.~P.~R.
  Minim}, \bibinfo{author}{A.~N. da~Silva}, \bibinfo{author}{A.~A. Simiqueli},
  \bibinfo{author}{S.~M. Della~Lucia}, \bibinfo{author}{L.~A. Minim}, et~al.,
\newblock \bibinfo{title}{Balanced incomplete block design: an alternative for
  data collection in the optimized descriptive profile},
\newblock \bibinfo{journal}{Food Research International} \bibinfo{volume}{64}
  (\bibinfo{year}{2014}) \bibinfo{pages}{289--297}.
\bibitem[{Basu et~al.(2014)Basu, Bagchi, and Ghosh}]{basu2014design}
\bibinfo{author}{M.~Basu}, \bibinfo{author}{S.~Bagchi}, \bibinfo{author}{D.~K.
  Ghosh},
\newblock \bibinfo{title}{Design of an efficient load balancing algorithm using
  the symmetric balanced incomplete block design},
\newblock \bibinfo{journal}{Information Sciences} \bibinfo{volume}{278}
  (\bibinfo{year}{2014}) \bibinfo{pages}{221--230}.
\bibitem[{Madsen et~al.(2014)Madsen, Jensen, Salmer{\'o}n, Karlsen, Langseth,
  and Nielsen}]{madsen2014new}
\bibinfo{author}{A.~L. Madsen}, \bibinfo{author}{F.~Jensen},
  \bibinfo{author}{A.~Salmer{\'o}n}, \bibinfo{author}{M.~Karlsen},
  \bibinfo{author}{H.~Langseth}, \bibinfo{author}{T.~D. Nielsen},
\newblock \bibinfo{title}{A new method for vertical parallelisation of tan
  learning based on balanced incomplete block designs},
\newblock in: \bibinfo{editor}{L.~C. van~der Gaag}, \bibinfo{editor}{A.~J.
  Feelders} (Eds.), \bibinfo{booktitle}{7th European Workshop on Probabilistic
  Graphical Models}, \bibinfo{publisher}{Springer International Publishing},
  \bibinfo{address}{Cham}, \bibinfo{year}{2014}, pp. \bibinfo{pages}{302--317}.
\bibitem[{Corneil and Mathon(1978)}]{Corneil+:algo-techniques-adm1978}
\bibinfo{author}{D.~G. Corneil}, \bibinfo{author}{R.~Mathon},
\newblock \bibinfo{title}{Algorithmic techniques for the generation and
  analysis of strongly regular graphs and other combinatorial configurations},
\newblock \bibinfo{journal}{Annals of Discrete Mathematics} \bibinfo{volume}{2}
  (\bibinfo{year}{1978}) \bibinfo{pages}{1--32}.
\bibitem[{Gibbons and \"Osterg{\aa}rd(1996)}]{gibbons-design-theory-1996}
\bibinfo{author}{P.~Gibbons}, \bibinfo{author}{P.~\"Osterg{\aa}rd},
\newblock \bibinfo{title}{Computational methods in design theory},
\newblock in: \bibinfo{editor}{C.~Colbourn}, \bibinfo{editor}{J.~Dinitz}
  (Eds.), \bibinfo{booktitle}{The CRC handbook of combinatorial designs},
  \bibinfo{publisher}{Boca Raton: CRC Press}, \bibinfo{year}{1996}, pp.
  \bibinfo{pages}{730--740}.
\bibitem[{Prestwich(2003)}]{DBLP:conf/cp2003/PrestwichCP03}
\bibinfo{author}{S.~Prestwich},
\newblock \bibinfo{title}{A local search algorithm for balanced incomplete
  block designs},
\newblock in: \bibinfo{editor}{F.~Rossi} (Ed.), \bibinfo{booktitle}{9th
  International Conference on Principles and Practices of Constraint
  Programming ({CP2003})}, volume \bibinfo{volume}{2833} of
  \textit{\bibinfo{series}{Lecture Notes in Computer Science}},
  \bibinfo{publisher}{Springer}, \bibinfo{year}{2003}, pp.
  \bibinfo{pages}{53--64}.
\bibitem[{Rodr\'{\i}guez et~al.(2009)Rodr\'{\i}guez, Cotta, and
  Fern\'{a}ndez-Leiva}]{RuedaCF09}
\bibinfo{author}{D.~Rodr\'{\i}guez}, \bibinfo{author}{C.~Cotta},
  \bibinfo{author}{A.~Fern\'{a}ndez-Leiva},
\newblock \bibinfo{title}{Finding balanced incomplete block designs with
  metaheuristics},
\newblock in: \bibinfo{booktitle}{9th European Conference Evolutionary
  Computation in Combinatorial Optimization -- EvoCOP 2009}, volume
  \bibinfo{volume}{5482} of \textit{\bibinfo{series}{Lecture Notes in Computer
  Science}}, \bibinfo{publisher}{Springer}, \bibinfo{address}{Berlin
  Heidelberg}, \bibinfo{year}{2009}, pp. \bibinfo{pages}{156--167}.
\bibitem[{Faghihi and Tat(2016)}]{faghihi2014varphi}
\bibinfo{author}{M.~R. Faghihi}, \bibinfo{author}{S.~Tat},
\newblock \bibinfo{title}{On $\phi$ p-optimality of incomplete block designs:
  An algorithm},
\newblock \bibinfo{journal}{Communications in Statistics - Simulation and
  Computation} \bibinfo{volume}{45} (\bibinfo{year}{2016})
  \bibinfo{pages}{758--769}.
\bibitem[{Rodr\'{\i}guez et~al.(2011)Rodr\'{\i}guez, Cotta, and
  Fern\'{a}ndez-Leiva}]{DBLP:journals/ijcopi/RuedaCL11}
\bibinfo{author}{D.~Rodr\'{\i}guez}, \bibinfo{author}{C.~Cotta},
  \bibinfo{author}{A.~Fern\'{a}ndez-Leiva},
\newblock \bibinfo{title}{A memetic algorithm for designing balanced incomplete
  blocks},
\newblock \bibinfo{journal}{International Journal of Combinatorial Optimization
  Problems and Informatics (IJCOPI)} \bibinfo{volume}{2} (\bibinfo{year}{2011})
  \bibinfo{pages}{14--22}.
\bibitem[{Benhamou(1994)}]{benhamou1994study}
\bibinfo{author}{B.~Benhamou},
\newblock \bibinfo{title}{Study of symmetry in constraint satisfaction
  problems},
\newblock in: \bibinfo{booktitle}{2nd Workshop on Principles and Practice of
  Constraint Programming, PPCP 94}, \bibinfo{organization}{DTIC Document}, pp.
  \bibinfo{pages}{246--254}.
\bibitem[{Backofen and Will(2002)}]{backofen2002excluding}
\bibinfo{author}{R.~Backofen}, \bibinfo{author}{S.~Will},
\newblock \bibinfo{title}{Excluding symmetries in constraint-based search},
\newblock \bibinfo{journal}{Constraints} \bibinfo{volume}{7}
  (\bibinfo{year}{2002}) \bibinfo{pages}{333--349}.
\bibitem[{Fahle et~al.(2001)Fahle, Schamberger, and Sellmann}]{fahle2001}
\bibinfo{author}{T.~Fahle}, \bibinfo{author}{S.~Schamberger},
  \bibinfo{author}{M.~Sellmann},
\newblock \bibinfo{title}{Symmetry breaking},
\newblock in: \bibinfo{editor}{W.~T.} (Ed.), \bibinfo{booktitle}{7th
  International Conference on the Principles and Practice of Constraint
  Programming - CP 2001}, volume \bibinfo{volume}{2239} of
  \textit{\bibinfo{series}{Lecture Notes in Computer Science}},
  \bibinfo{publisher}{Springer-Verlag}, \bibinfo{year}{2001}, pp.
  \bibinfo{pages}{93--107}.
\bibitem[{Gent and Smith(1999)}]{Gent99symmetrybreaking}
\bibinfo{author}{I.~P. Gent}, \bibinfo{author}{B.~Smith},
\newblock \bibinfo{title}{Symmetry breaking in constraint programming},
\newblock in: \bibinfo{editor}{W.~Horn} (Ed.), \bibinfo{booktitle}{14th
  European Conference on Artificial Intelligence -- ECAI 2000},
  \bibinfo{publisher}{{IOS} Press}, \bibinfo{year}{1999}, pp.
  \bibinfo{pages}{599--603}.
\bibitem[{Meseguer and Torras(2001)}]{DBLP:journals/ai/MeseguerT01}
\bibinfo{author}{P.~Meseguer}, \bibinfo{author}{C.~Torras},
\newblock \bibinfo{title}{Exploiting symmetries within constraint satisfaction
  search},
\newblock \bibinfo{journal}{Artif. Intell.} \bibinfo{volume}{129}
  (\bibinfo{year}{2001}) \bibinfo{pages}{133--163}.
\bibitem[{Cochran and Cox(1957)}]{cox}
\bibinfo{author}{W.~G. Cochran}, \bibinfo{author}{G.~M. Cox},
  \bibinfo{title}{Experimental Design}, \bibinfo{publisher}{John Wiley},
  \bibinfo{address}{New York}, \bibinfo{year}{1957}.
\bibitem[{Fisher and Yates(1949)}]{yates2}
\bibinfo{author}{R.~A. Fisher}, \bibinfo{author}{F.~Yates},
  \bibinfo{title}{Statistical Tables for Biological, Agricultural and Medical
  Research}, \bibinfo{publisher}{Oliver \& Boy}, \bibinfo{edition}{3} edition,
  \bibinfo{year}{1949}.
\bibitem[{Whitaker et~al.(1990)Whitaker, Triggs, and John}]{WhitakerTriggs1990}
\bibinfo{author}{D.~Whitaker}, \bibinfo{author}{C.~M. Triggs},
  \bibinfo{author}{J.~A. John},
\newblock \bibinfo{title}{Construction of block designs using mathematical
  programming},
\newblock \bibinfo{journal}{Journal of the Royal Statistical Society. Series B
  (Methodological)} \bibinfo{volume}{52} (\bibinfo{year}{1990})
  \bibinfo{pages}{497--503}.
\bibitem[{Zergaw(1989)}]{zergaw}
\bibinfo{author}{G.~Zergaw},
\newblock \bibinfo{title}{A sequential method of constructing optimal block
  designs},
\newblock \bibinfo{journal}{Australian \& New Zealand Journal of Statistics}
  \bibinfo{volume}{31} (\bibinfo{year}{1989}) \bibinfo{pages}{333--342}.
\bibitem[{Tjur(1993)}]{tjur}
\bibinfo{author}{T.~Tjur},
\newblock \bibinfo{title}{An algorithm for optimization of block designs},
\newblock \bibinfo{journal}{Journal of Statistical Planning and Inference}
  \bibinfo{volume}{36} (\bibinfo{year}{1993}) \bibinfo{pages}{277--282}.
\bibitem[{Flener et~al.(2001)Flener, Frisch, Hnich, Kzltan, Miguel, and
  Walsh}]{flener01}
\bibinfo{author}{P.~Flener}, \bibinfo{author}{A.~M. Frisch},
  \bibinfo{author}{B.~Hnich}, \bibinfo{author}{Z.~Kzltan},
  \bibinfo{author}{I.~Miguel}, \bibinfo{author}{T.~Walsh},
\newblock \bibinfo{title}{Matrix modelling},
\newblock in: \bibinfo{editor}{B.~Smith}, \bibinfo{editor}{K.~Brown},
  \bibinfo{editor}{P.~Prosser}, \bibinfo{editor}{I.~P. Gent} (Eds.),
  \bibinfo{booktitle}{CP-01 Workshop on Modelling and Problem Formulation.
  International Conference on the Principles and Practice of Constraint
  Programming}, \bibinfo{address}{Paphos, Cyprus}, pp. \bibinfo{pages}{1--7}.
\bibitem[{Puget(2002)}]{DBLP:conf/cp/Puget02}
\bibinfo{author}{J.-F. Puget},
\newblock \bibinfo{title}{Symmetry breaking revisited},
\newblock in: \bibinfo{editor}{P.~V. Hentenryck} (Ed.), \bibinfo{booktitle}{8th
  International Conference on the Principles and Practice of Constraint
  Programming ({CP 2002})}, volume \bibinfo{volume}{2470} of
  \textit{\bibinfo{series}{Lecture Notes in Computer Science}},
  \bibinfo{publisher}{Springer}, \bibinfo{address}{Ithaca, NY, USA},
  \bibinfo{year}{2002}, pp. \bibinfo{pages}{446--461}.
\bibitem[{Bofill et~al.(2003)Bofill, Guimer{\`a}, and
  Torras}]{DBLP:journals/nn/BofillGT03}
\bibinfo{author}{P.~Bofill}, \bibinfo{author}{R.~Guimer{\`a}},
  \bibinfo{author}{C.~Torras},
\newblock \bibinfo{title}{Comparison of simulated annealing and mean field
  annealing as applied to the generation of block designs},
\newblock \bibinfo{journal}{Neural Networks} \bibinfo{volume}{16}
  (\bibinfo{year}{2003}) \bibinfo{pages}{1421--1428}.
\bibitem[{Prestwich(2003)}]{prestwich:negative-efefcts-aor03}
\bibinfo{author}{S.~Prestwich},
\newblock \bibinfo{title}{Negative effects of modeling techniques on search
  performance},
\newblock \bibinfo{journal}{Annals of Operations Research} \bibinfo{volume}{18}
  (\bibinfo{year}{2003}) \bibinfo{pages}{137--150}.
\bibitem[{Yokoya and Yamada(2011)}]{yokoya2011mathematical}
\bibinfo{author}{D.~Yokoya}, \bibinfo{author}{T.~Yamada},
\newblock \bibinfo{title}{A mathematical programming approach to the
  construction of bibds},
\newblock \bibinfo{journal}{International Journal of Computer Mathematics}
  \bibinfo{volume}{88} (\bibinfo{year}{2011}) \bibinfo{pages}{1067--1082}.
\bibitem[{Mandal(2015)}]{Mandal2014183}
\bibinfo{author}{B.~N. Mandal},
\newblock \bibinfo{title}{Linear integer programming approach to construction
  of balanced incomplete block designs},
\newblock \bibinfo{journal}{Communications in Statistics - Simulation and
  Computation} \bibinfo{volume}{44} (\bibinfo{year}{2015})
  \bibinfo{pages}{1405--1411}.
\bibitem[{Rodr\'{\i}guez et~al.(2016)Rodr\'{\i}guez, Darghan, and
  Monroy}]{david:jorunal_colombian_2016}
\bibinfo{author}{D.~Rodr\'{\i}guez}, \bibinfo{author}{E.~Darghan},
  \bibinfo{author}{J.~Monroy},
\newblock \bibinfo{title}{A multi-agent proposal in the resolution of instances
  of {BIBD}},
\newblock \bibinfo{journal}{Revista Colombiana de Estad\'{\i}stica}
  \bibinfo{volume}{39} (\bibinfo{year}{2016}) \bibinfo{pages}{267--280}.
\bibitem[{Neri et~al.(2012)Neri, Cotta, and
  Moscato}]{HandbookMemeticNeriCotta2012}
\bibinfo{author}{F.~Neri}, \bibinfo{author}{C.~Cotta},
  \bibinfo{author}{P.~Moscato}, \bibinfo{title}{Handbook of Memetic
  Algorithms}, volume \bibinfo{volume}{379} of \textit{\bibinfo{series}{Studies
  in Computational Intelligence}}, \bibinfo{publisher}{Springer Berlin
  Heidelberg}, \bibinfo{year}{2012}.
\bibitem[{Rothlauf(2006)}]{DBLP:books/daglib/0014740}
\bibinfo{author}{F.~Rothlauf}, \bibinfo{title}{Representations for genetic and
  evolutionary algorithms}, \bibinfo{publisher}{Springer}, \bibinfo{edition}{2}
  edition, \bibinfo{year}{2006}.
\bibitem[{Camp\^{e}lo et~al.(2008)Camp\^{e}lo, Campos, and
  Corr\^{e}a}]{CAMPELO20081097}
\bibinfo{author}{M.~Camp\^{e}lo}, \bibinfo{author}{V.~A. Campos},
  \bibinfo{author}{R.~C. Corr\^{e}a},
\newblock \bibinfo{title}{On the asymmetric representatives formulation for the
  vertex coloring problem},
\newblock \bibinfo{journal}{Discrete Applied Mathematics} \bibinfo{volume}{156}
  (\bibinfo{year}{2008}) \bibinfo{pages}{1097 -- 1111}. \bibinfo{note}{GRACO
  2005}.
\bibitem[{Jans and Desrosiers(2010)}]{G201044}
\bibinfo{author}{R.~Jans}, \bibinfo{author}{J.~Desrosiers},
\newblock \bibinfo{title}{Binary clustering problems: Symmetric, asymmetric and
  decomposition formulations},
\newblock \bibinfo{journal}{GERAD Technical Report G-2010-44}
  (\bibinfo{year}{2010}) \bibinfo{pages}{1 -- 15}.
\bibitem[{Jans and Desrosiers(2013)}]{JANS20131132}
\bibinfo{author}{R.~Jans}, \bibinfo{author}{J.~Desrosiers},
\newblock \bibinfo{title}{Efficient symmetry breaking formulations for the job
  grouping problem},
\newblock \bibinfo{journal}{Computers \& Operations Research}
  \bibinfo{volume}{40} (\bibinfo{year}{2013}) \bibinfo{pages}{1132 -- 1142}.
\bibitem[{Vo-Thanh et~al.(2018)Vo-Thanh, Jans, Schoen, and Goos}]{G2016117}
\bibinfo{author}{N.~Vo-Thanh}, \bibinfo{author}{R.~Jans},
  \bibinfo{author}{E.~D. Schoen}, \bibinfo{author}{P.~Goos},
\newblock \bibinfo{title}{Symmetry breaking in mixed integer linear programming
  formulations for blocking two-level orthogonal experimental designs},
\newblock \bibinfo{journal}{Computers \& Operations Research}
  \bibinfo{volume}{97} (\bibinfo{year}{2018}) \bibinfo{pages}{96 -- 110}.
\bibitem[{Puchinger and Raidl(2005)}]{Puchinger2005}
\bibinfo{author}{J.~Puchinger}, \bibinfo{author}{G.~R. Raidl},
\newblock \bibinfo{title}{Combining metaheuristics and exact algorithms in
  combinatorial optimization: A survey and classification},
\newblock in: \bibinfo{editor}{J.~Mira}, \bibinfo{editor}{J.~\'{A}lvarez}
  (Eds.), \bibinfo{booktitle}{Artificial Intelligence and Knowledge Engineering
  Applications: A Bioinspired Approach}, volume \bibinfo{volume}{3562} of
  \textit{\bibinfo{series}{Lecture Notes in Computer Science}},
  \bibinfo{publisher}{Springer}, \bibinfo{address}{Berlin Heidelberg},
  \bibinfo{year}{2005}, pp. \bibinfo{pages}{113--124}.
\bibitem[{Crainic and Toulouse(2008)}]{Crainic2008}
\bibinfo{author}{T.~G. Crainic}, \bibinfo{author}{M.~Toulouse},
\newblock \bibinfo{title}{Explicit and emergent cooperation schemes for search
  algorithms},
\newblock in: \bibinfo{editor}{V.~Maniezzo}, \bibinfo{editor}{R.~Battiti},
  \bibinfo{editor}{J.-P. Watson} (Eds.), \bibinfo{booktitle}{Learning and
  Intelligent Optimization -- LION 2007 II}, \bibinfo{publisher}{Springer},
  \bibinfo{address}{Berlin Heidelberg}, \bibinfo{year}{2008}, pp.
  \bibinfo{pages}{95--109}.
\bibitem[{Cruz and Pelta(2009)}]{Cruz2009}
\bibinfo{author}{C.~Cruz}, \bibinfo{author}{D.~Pelta},
\newblock \bibinfo{title}{Soft computing and cooperative strategies for
  optimization},
\newblock \bibinfo{journal}{Applied Soft Computing} \bibinfo{volume}{9}
  (\bibinfo{year}{2009}) \bibinfo{pages}{30--38}.
\bibitem[{Masegosa et~al.(2009)Masegosa, Mascia, Pelta, and
  Brunato}]{Masegosa2009}
\bibinfo{author}{A.~Masegosa}, \bibinfo{author}{F.~Mascia},
  \bibinfo{author}{D.~Pelta}, \bibinfo{author}{M.~Brunato},
\newblock \bibinfo{title}{Cooperative strategies and reactive search: A hybrid
  model proposal},
\newblock in: \bibinfo{editor}{T.~St\"{u}tzle} (Ed.),
  \bibinfo{booktitle}{Learning and Intelligent Optimization}, volume
  \bibinfo{volume}{5851} of \textit{\bibinfo{series}{Lecture Notes in Computer
  Science}}, \bibinfo{publisher}{Springer}, \bibinfo{address}{Berlin
  Heidelberg}, \bibinfo{year}{2009}, pp. \bibinfo{pages}{206--220}.
\bibitem[{Amaya et~al.(2011{\natexlab{a}})Amaya, Cotta, and
  Fern\'{a}ndez-Leiva}]{Amaya2011}
\bibinfo{author}{J.~Amaya}, \bibinfo{author}{C.~Cotta},
  \bibinfo{author}{A.~Fern\'{a}ndez-Leiva},
\newblock \bibinfo{title}{A memetic cooperative optimization schema and its
  application to the tool switching problem},
\newblock in: \bibinfo{editor}{R.~Schaefer}, et~al. (Eds.),
  \bibinfo{booktitle}{Parallel Problem Solving from Nature - PPSN XI}, volume
  \bibinfo{volume}{6238} of \textit{\bibinfo{series}{Lecture Notes in Computer
  Science}}, \bibinfo{publisher}{Springer}, \bibinfo{address}{Berlin
  Heidelberg}, \bibinfo{year}{2011}{\natexlab{a}}, pp.
  \bibinfo{pages}{445--454}.
\bibitem[{Amaya et~al.(2011{\natexlab{b}})Amaya, Cotta, and
  Fern\'{a}ndez-Leiva}]{Amaya2011MC}
\bibinfo{author}{J.~Amaya}, \bibinfo{author}{C.~Cotta},
  \bibinfo{author}{A.~Fern\'{a}ndez-Leiva},
\newblock \bibinfo{title}{Memetic cooperative models for the tool switching
  problem},
\newblock \bibinfo{journal}{Memetic Computing} \bibinfo{volume}{3}
  (\bibinfo{year}{2011}{\natexlab{b}}) \bibinfo{pages}{199--216}.
\bibitem[{Nogueras and Cotta(2014)}]{NoguerasCotta2014}
\bibinfo{author}{R.~Nogueras}, \bibinfo{author}{C.~Cotta},
\newblock \bibinfo{title}{An analysis of migration strategies in island-based
  multimemetic algorithms},
\newblock in: \bibinfo{editor}{T.~Bartz-Beielstein},
  \bibinfo{editor}{J.~Branke}, \bibinfo{editor}{B.~Filipi{\'c}},
  \bibinfo{editor}{J.~Smith} (Eds.), \bibinfo{booktitle}{Parallel Problem
  Solving from Nature -- PPSN XIII}, volume \bibinfo{volume}{8672} of
  \textit{\bibinfo{series}{Lecture Notes in Computer Science}},
  \bibinfo{publisher}{Springer}, \bibinfo{address}{Berlin Heidelberg},
  \bibinfo{year}{2014}, pp. \bibinfo{pages}{731--740}.
\bibitem[{Friedman(1937)}]{Friedman1937}
\bibinfo{author}{M.~Friedman},
\newblock \bibinfo{title}{The use of ranks to avoid the assumption of normality
  implicit in the analysis of variance},
\newblock \bibinfo{journal}{Journal of the American Statistical Association}
  \bibinfo{volume}{32} (\bibinfo{year}{1937}) \bibinfo{pages}{675--701}.
\bibitem[{Iman and Davenport(1980)}]{Iman1980}
\bibinfo{author}{R.~Iman}, \bibinfo{author}{J.~Davenport},
\newblock \bibinfo{title}{Approximations of the critical region of the
  {F}riedman statistic},
\newblock \bibinfo{journal}{Communications in Statistics} \bibinfo{volume}{9}
  (\bibinfo{year}{1980}) \bibinfo{pages}{571--595}.
\bibitem[{Holm(1979)}]{Holm1979}
\bibinfo{author}{S.~Holm},
\newblock \bibinfo{title}{A simple sequentially rejective multiple test
  procedure},
\newblock \bibinfo{journal}{Scandinavian Journal of Statistics}
  \bibinfo{volume}{6} (\bibinfo{year}{1979}) \bibinfo{pages}{65--70}.
\bibitem[{Lehmann and D'Abrera(1998)}]{Lehman1998}
\bibinfo{author}{E.~Lehmann}, \bibinfo{author}{H.~D'Abrera},
  \bibinfo{title}{Nonparametrics: statistical methods based on ranks},
  \bibinfo{publisher}{Prentice-Hall}, \bibinfo{address}{Englewood Cliffs, NJ},
  \bibinfo{year}{1998}.
\bibitem[{Rodr\'{\i}guez et~al.(2017)Rodr\'{\i}guez, Cotta, and
  Fern\'{a}ndez-Leiva}]{datainbriefSubmitted}
\bibinfo{author}{D.~Rodr\'{\i}guez}, \bibinfo{author}{C.~Cotta},
  \bibinfo{author}{A.~Fern\'{a}ndez-Leiva},
\newblock \bibinfo{title}{Data for the search of 2-designs: solutions and
  metaheuristics code},
\newblock \bibinfo{journal}{Data in Brief}  (\bibinfo{year}{2017}).
  \bibinfo{note}{Submitted}.
\bibitem[{Smith(2012)}]{Smith2012}
\bibinfo{author}{J.~Smith},
\newblock \bibinfo{title}{Self-adaptative and coevolving memetic algorithms},
\newblock in: \bibinfo{editor}{F.~Neri}, \bibinfo{editor}{C.~Cotta},
  \bibinfo{editor}{P.~Moscato} (Eds.), \bibinfo{booktitle}{Handbook of Memetic
  Algorithms}, volume \bibinfo{volume}{379} of \textit{\bibinfo{series}{Studies
  in Computational Intelligence}}, \bibinfo{publisher}{Springer Berlin
  Heidelberg}, \bibinfo{year}{2012}, pp. \bibinfo{pages}{167--188}.

\end{thebibliography}
\end{document}